\documentclass[10pt,twocolumn,letterpaper]{article}
\usepackage{cvpr}
\usepackage{times}
\usepackage{epsfig}
\usepackage{graphicx}
\usepackage{amsmath}
\usepackage{amssymb}
\usepackage{subcaption}
\usepackage{epsfig}
\usepackage{booktabs}
\usepackage{tabularx}
\usepackage[english]{babel}
\usepackage[utf8]{inputenc}
\usepackage{algorithm}
\usepackage[noend]{algpseudocode}
\makeatletter

\usepackage[pagebackref=true,breaklinks=true,letterpaper=true,colorlinks,bookmarks=false]{hyperref}

\cvprfinalcopy 

\def\cvprPaperID{1944} 

\ifcvprfinal\fi
\begin{document}

\title{DeblurGAN: Blind Motion Deblurring Using Conditional Adversarial Networks}

\author{Orest Kupyn\textsuperscript{1,3}, Volodymyr  Budzan\textsuperscript{1,3}, Mykola  Mykhailych\textsuperscript{1}, Dmytro Mishkin\textsuperscript{2}, Ji\v{r}i Matas\textsuperscript{2} \vspace{2mm} \\
  \textsuperscript{1} Ukrainian Catholic University, Lviv, Ukraine\\
  \texttt{\{kupyn, budzan, mykhailych\}@ucu.edu.ua} \vspace{2mm} \\
  \textsuperscript{2} Visual Recognition Group, Center for Machine Perception, FEE, CTU in Prague \\
  \texttt{\{mishkdmy,  matas\}@cmp.felk.cvut.cz}\\
  \textsuperscript{3} ELEKS Ltd.\\
  }
\maketitle
\begin{abstract}
We present DeblurGAN, an end-to-end learned method for motion deblurring. The learning is based on a conditional GAN and the content loss . DeblurGAN achieves state-of-the art performance   both in the structural similarity measure and visual appearance.
The quality of the deblurring model is also evaluated in a novel way on a real-world problem -- object detection on (de-)blurred images.  
The method is 5 times faster than the closest competitor -- DeepDeblur~\cite{Nah2016DeepDeblurring}. We also introduce a novel method for generating synthetic motion blurred images from  sharp ones, allowing realistic dataset augmentation.  

The model, code and the  dataset are available at \url{https://github.com/KupynOrest/DeblurGAN}
\end{abstract}
\vspace{-2em}
\section{Introduction}
\label{s:intro}
\begin{figure}[htb]
\centering
 \includegraphics[width=0.95\linewidth]{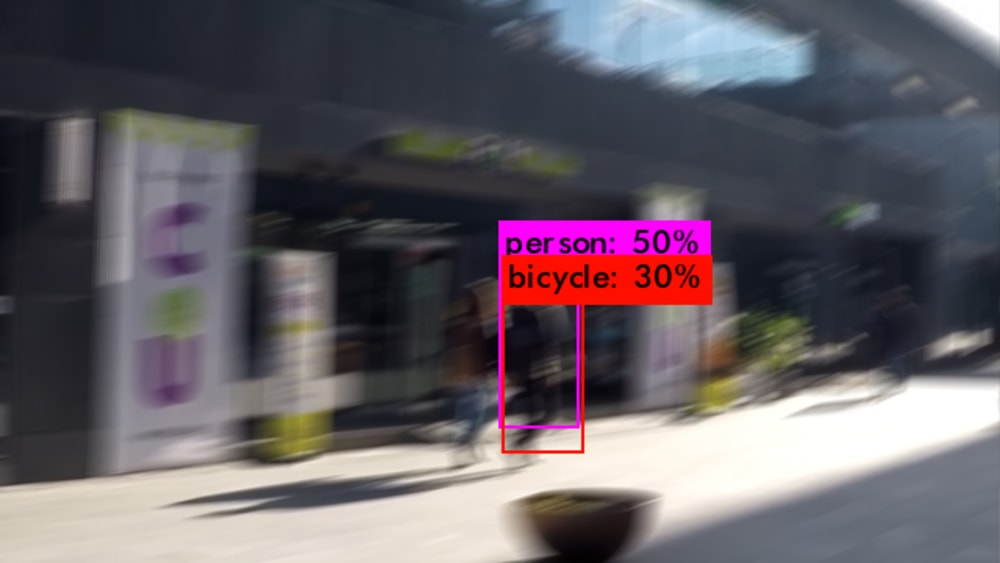}\\
  \includegraphics[width=0.95\linewidth]{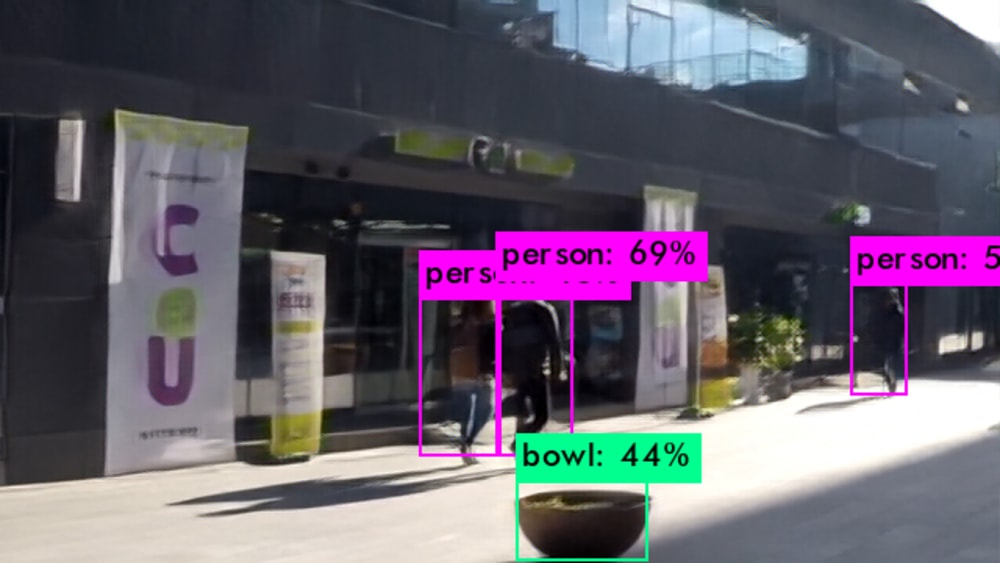} \\
  \includegraphics[width=0.95\linewidth]{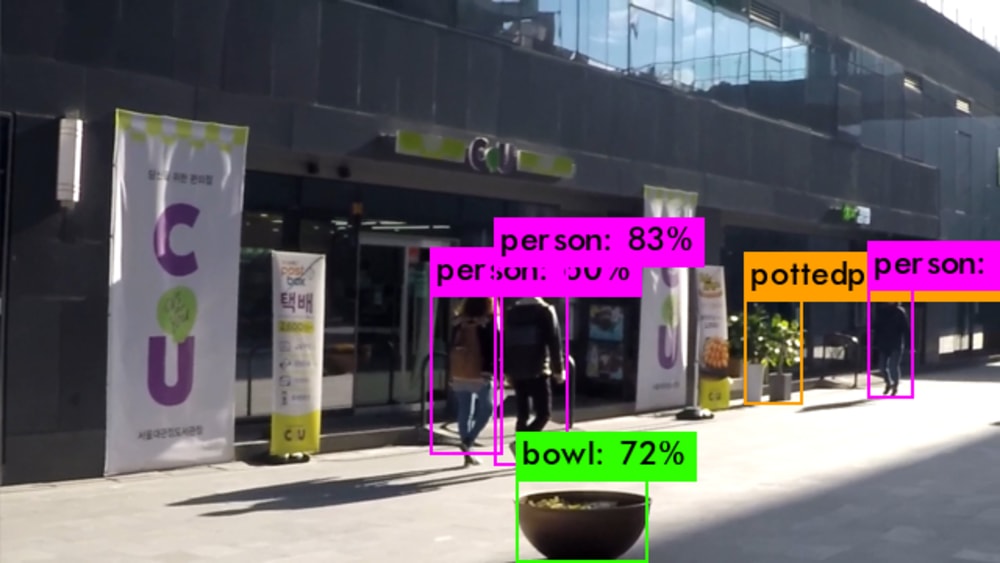}\\
   \caption{DeblurGAN helps object detection. YOLO~\cite{YOLO} detections on the blurred image (top), the DeblurGAN restored (middle) and the sharp ground truth image from the GoPro~\cite{Nah2016DeepDeblurring} dataset.
   \vspace{-2em}}
 \label{fig:cover}
\end{figure}
\newcommand{\ra}[1]{\renewcommand{\arraystretch}{#1}}
    This work is on blind motion deblurring of a single photograph. Significant progress has been recently achieved in related areas of image super-resolution~\cite{SRGAN} and in-painting~\cite{Inpainting} by applying generative adversarial networks (GANs)~\cite{GAN}. GANs are known for the ability to preserve texture details in images, create solutions that are close to the real image manifold and look perceptually convincing.
        \begin{figure*}[tb]
	\centering
        \includegraphics[width=0.33\textwidth]{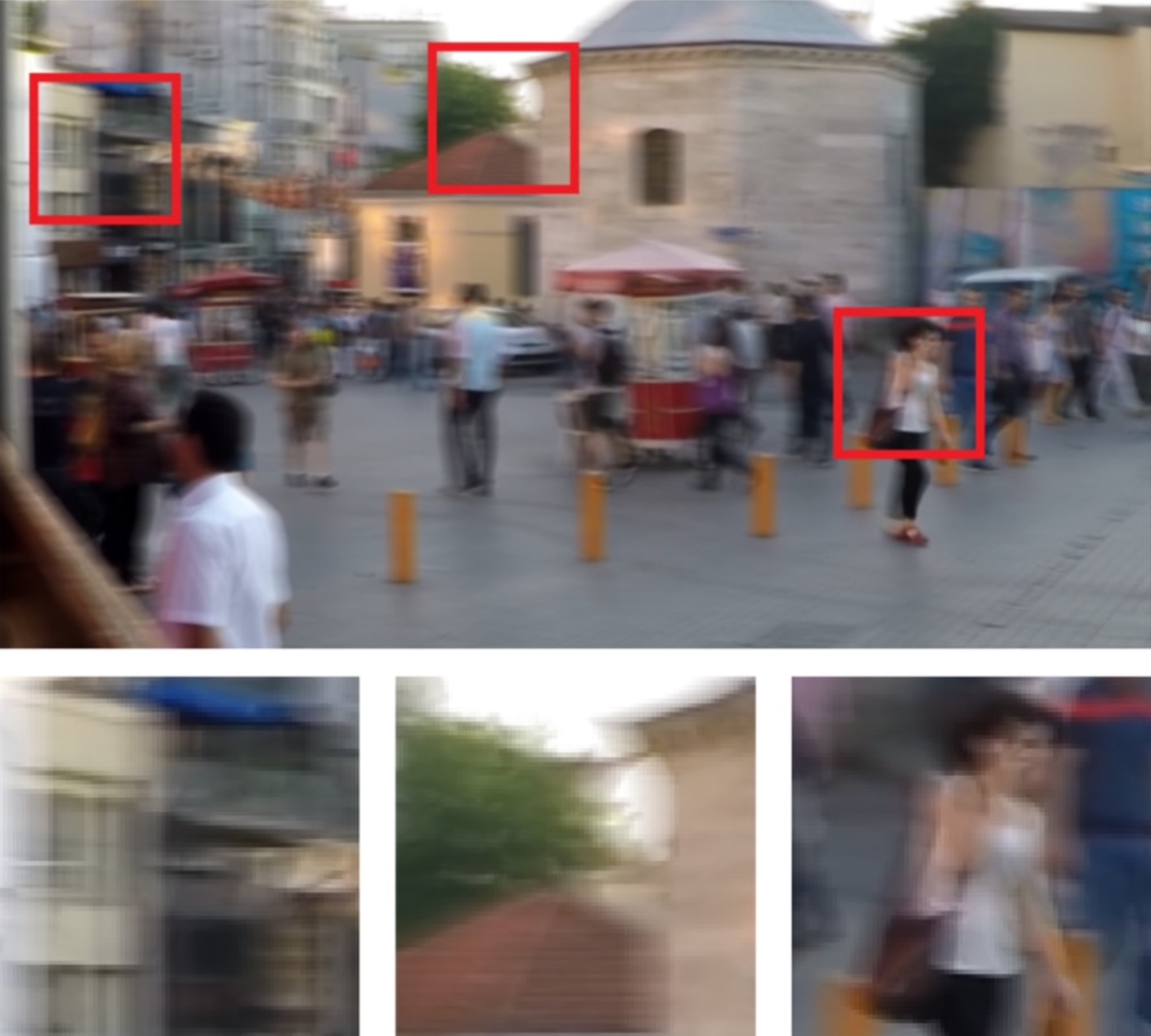}
        \includegraphics[width=0.33\textwidth]{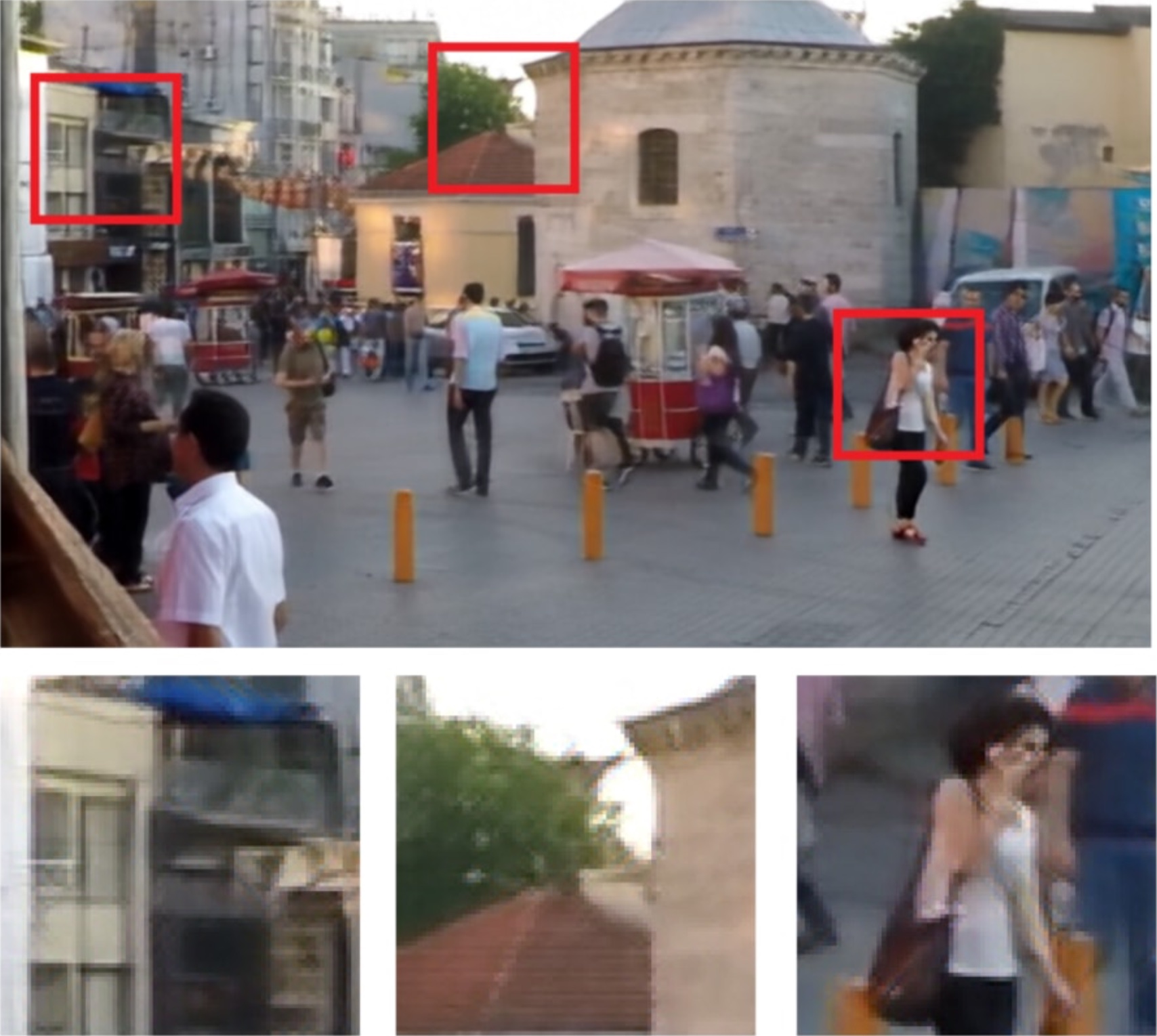}
        \includegraphics[width=0.33\textwidth]{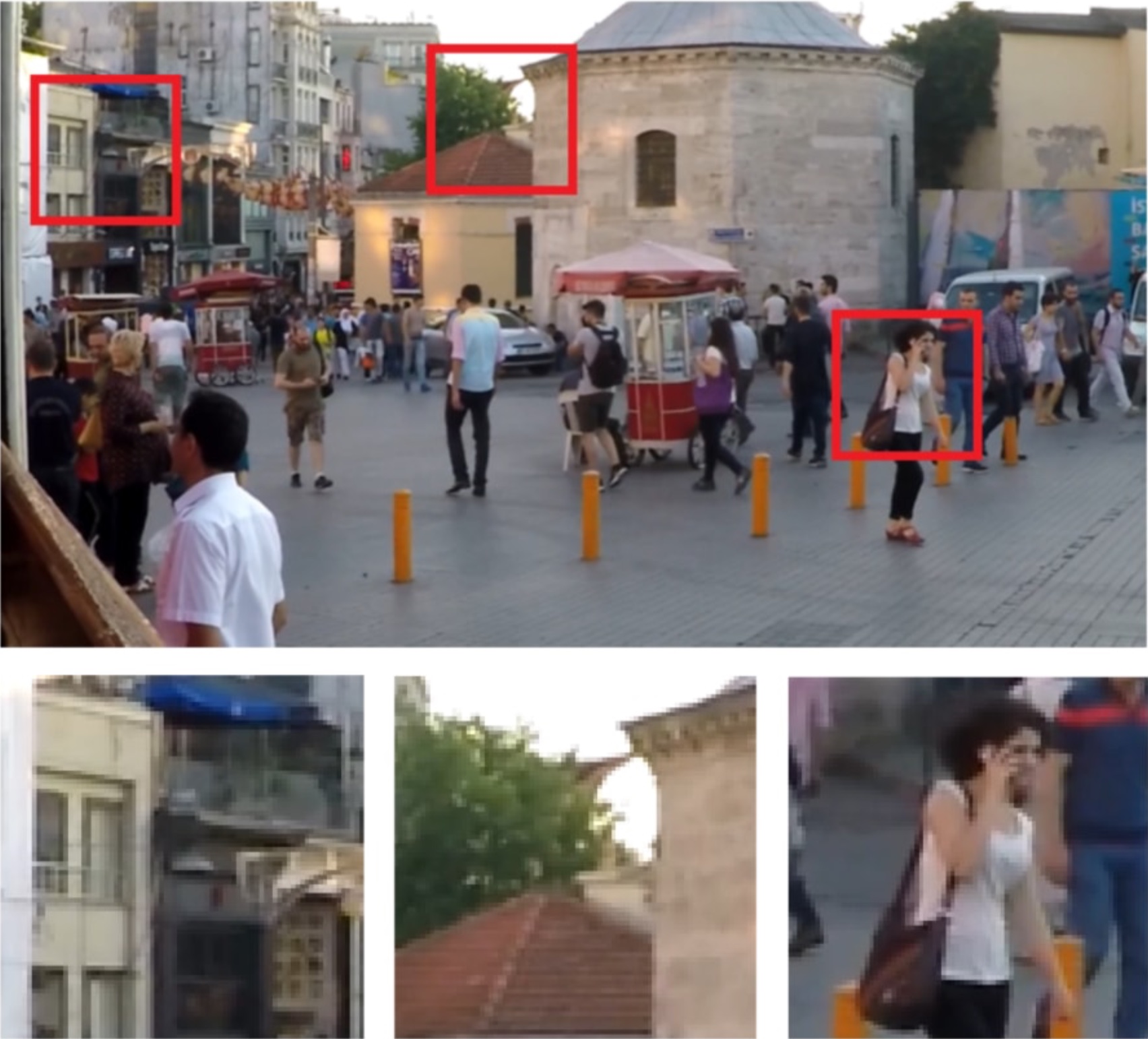}\\
        \vspace{0.2em}
        \includegraphics[width=0.33\textwidth]{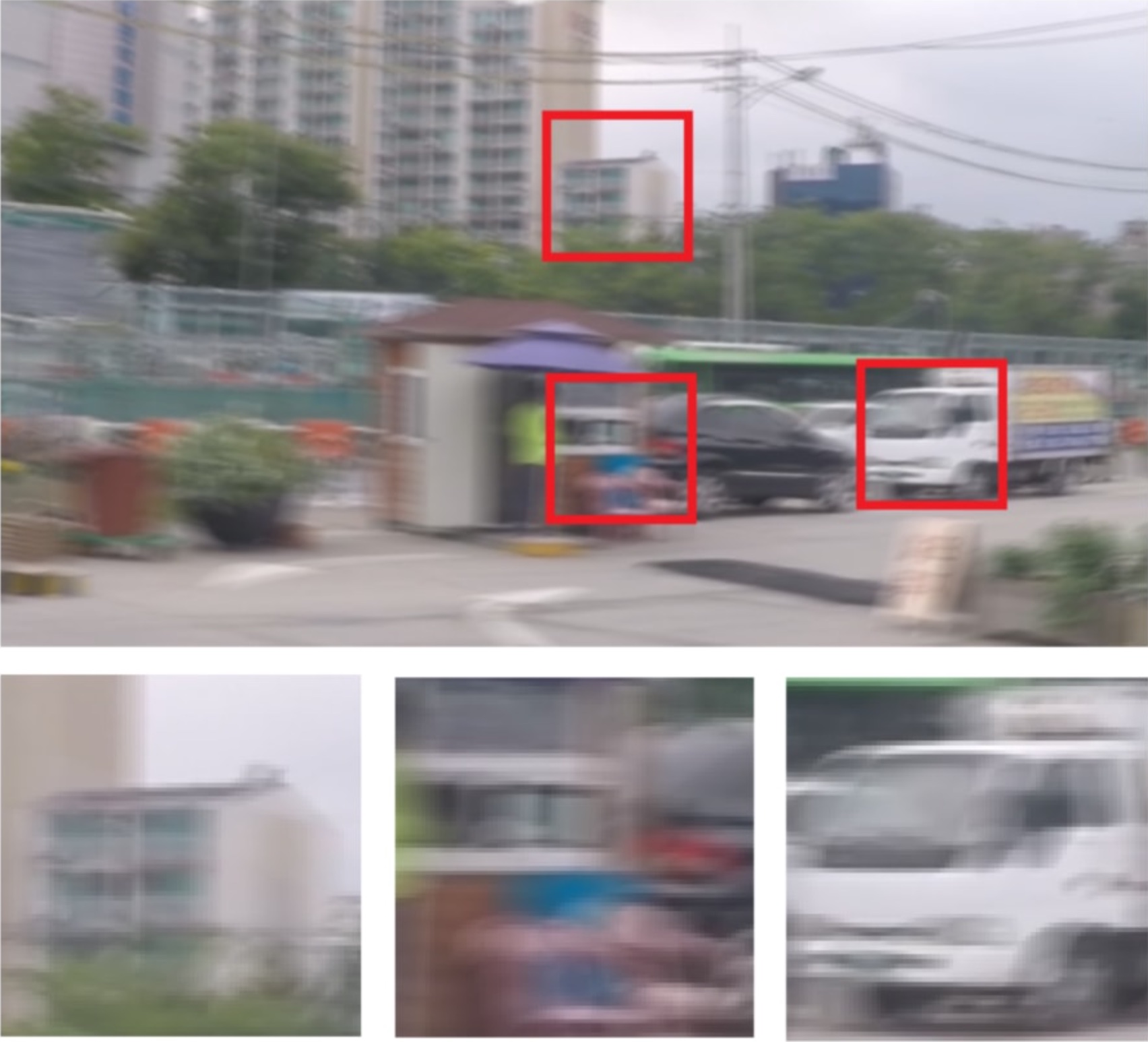}
        \includegraphics[width=0.33\textwidth]{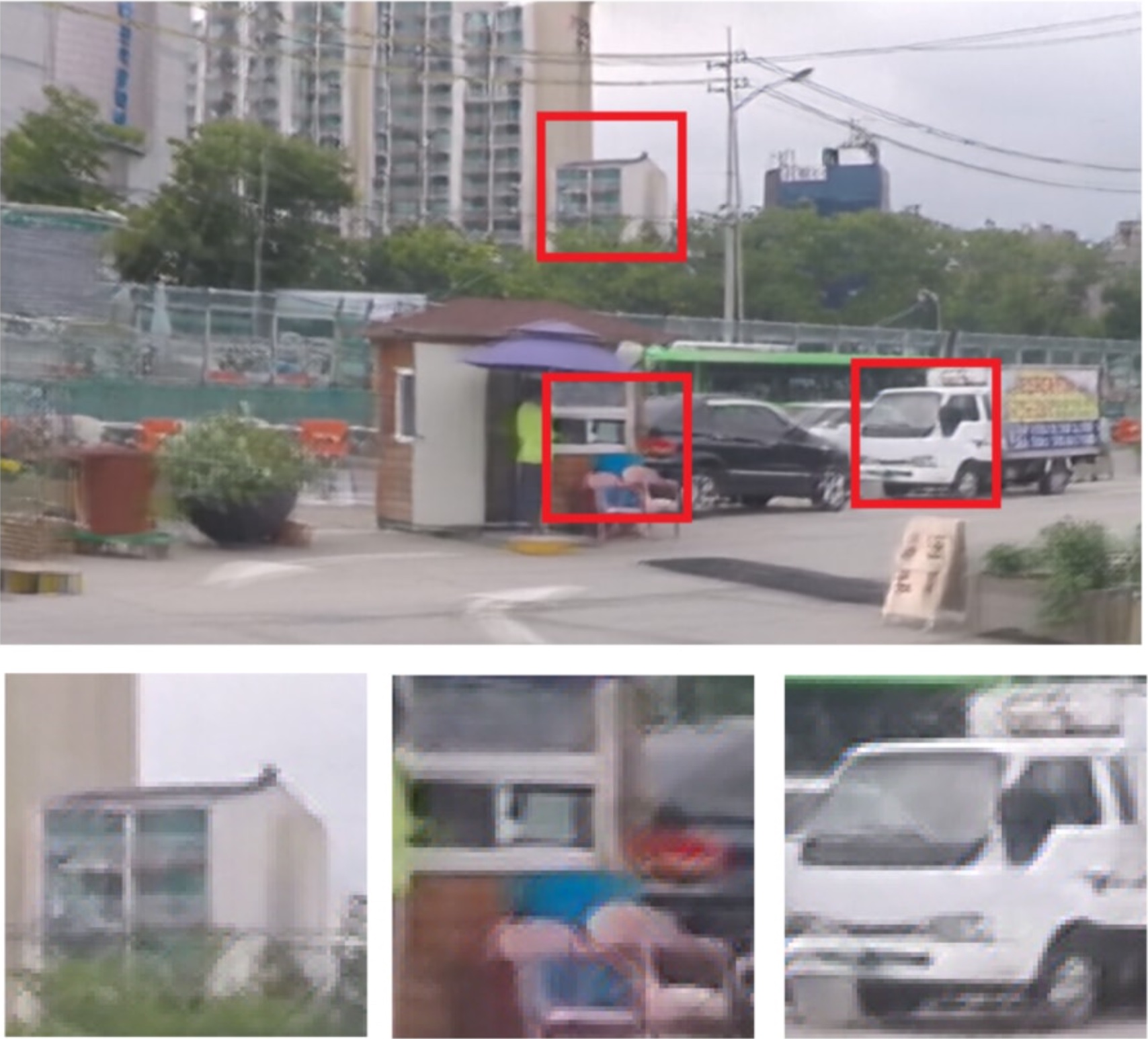}
        \includegraphics[width=0.33\textwidth]{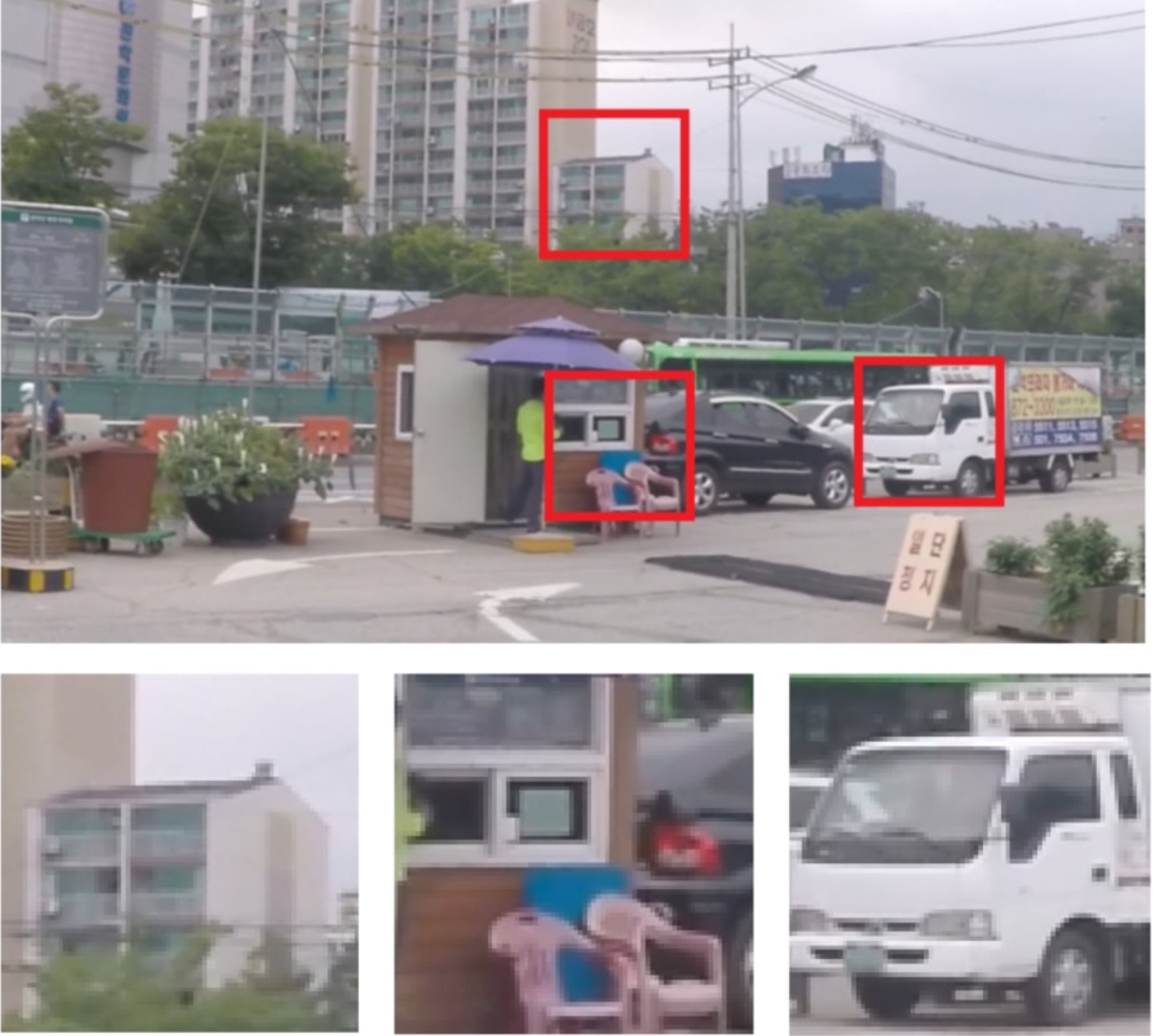}
    \caption{GoPro images~\cite{Nah2016DeepDeblurring} processed by DeblurGAN. Blurred -- left, DeblurGAN -- center, ground truth sharp -- right.}\label{fig:our_model}
\end{figure*}
    Inspired by recent work on image super-resolution~\cite{SRGAN} and image-to-image translation by generative adversarial networks~\cite{pix2pix}, we treat deblurring as a special case of such image-to-image translation. 
We present DeblurGAN -- an approach based on conditional generative adversarial networks~\cite{CGAN} and a multi-component loss function. Unlike previous work we use Wasserstein GAN~\cite{WGAN} with the gradient penalty~\cite{WGAN-GP} and perceptual loss~\cite{Johnson2016Perceptual}. This encourages solutions which are perceptually hard to distinguish from real sharp images and allows to restore finer texture details than if using traditional MSE or MAE as an optimization target. 
  
We make three contributions. First, we propose a loss and architecture which obtain state-of-the art results in motion deblurring, while being 5x faster than the fastest competitor. Second, we present a method based on random trajectories for generating a dataset for motion deblurring training in an automated fashion from the set of sharp image. We show that combining it with an existing dataset for motion deblurring learning improves results compared to training on real-world images only.
Finally, we present a novel dataset and method for evaluation of deblurring algorithms based on how they improve object detection results.
\section{Related work}
\label{s:related-work}
\subsection{Image Deblurring}
The common formulation of non-uniform blur model is the following:
    \begin{equation}
    \label{eq:2}
    I_B = k(M) * I_S + N,
    \end{equation}
where $I_B$ is a blurred image, $k(M)$ are unknown blur kernels determined by motion field $M$.
$I_S$ is the sharp latent image, $*$ denotes the convolution, $N$ is an additive noise. \\
\begin{figure*}[htb]
  \includegraphics[width=0.99\textwidth]{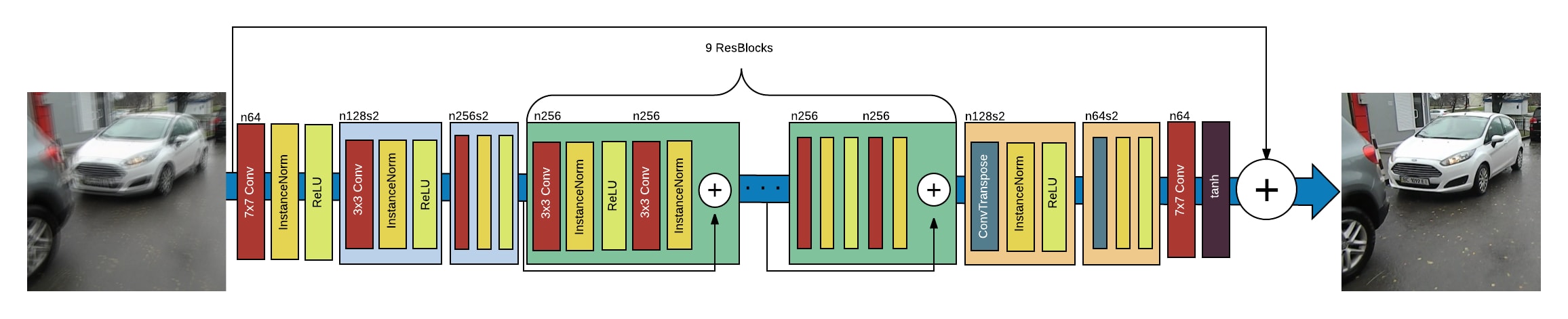}
  \caption{DeblurGAN generator architecture. DeblurGAN contains two strided convolution blocks with stride $\frac{1}{2}$, nine residual blocks~\cite{ResNet} and two transposed convolution blocks. Each ResBlock consists of a convolution layer, instance normalization layer, and ReLU activation.}
  \label{fig:arch}
\end{figure*}
    The family of deblurring problems is divided into two types: blind and non-blind deblurring. Early work~\cite{Szeliski} mostly focused on non-blind deblurring, making an assumption that the blur kernels $k(M)$ are known. Most rely on the classical Lucy-Richardson algorithm, Wiener or Tikhonov filter to perform the deconvolution operation and obtain $I_S$ estimate. Commonly the blur function is unknown, and blind deblurring algorithms estimate both latent sharp image $I_S$ and blur kernels $k(M)$. Finding a blur function for each pixel is an ill-posed problem, and most of the existing algorithms rely on heuristics, image statistics and assumptions on the sources of the blur.
    Those family of methods addresses the blur caused by camera shake by considering blur to be uniform across the image. Firstly, the camera motion is estimated in terms of the induced blur kernel, and then the effect is reversed by performing a deconvolution operation. Starting with the success of Fergus~\etal\cite{Fergus}, many methods~\cite{XuUnnaturalDeblurring}\cite{Xu2010}\cite{Fergus++}\cite{Fergus++1} has been developed over the last ten years. Some of the methods are based on an iterative approach~\cite{Fergus}~\cite{XuUnnaturalDeblurring}, which improve the estimate of the motion kernel and sharp image on each iteration by using parametric prior models. However, the running time, as well as the stopping criterion, is a significant problem for those kinds of algorithms. Others use assumptions of a local linearity of a blur function and simple heuristics to quickly estimate the unknown kernel. These methods are fast but work well on a small subset of images.
    
    Recently, Whyte~\etal~\cite{whyte} developed a novel algorithm for non-uniform blind deblurring based on a parametrized geometric model of the blurring process in terms of the rotational velocity of the camera during exposure. Similarly Gupta~\etal~\cite{3DDeblur} made an assumption that the blur is caused only by 3D camera movement.
	With the success of deep learning, over the last few years, there appeared some approaches based on convolutional neural networks (CNNs). Sun~\etal~\cite{SunLearningRemoval} use CNN to estimate blur kernel, Chakrabarti~\cite{Chakrabarti2016ADeblurring} predicts complex Fourier coefficients of motion kernel to perform non-blind deblurring in Fourier space whereas Gong~\cite{GongFromBlur} use fully convolutional network to move for motion flow estimation. All of these approaches use CNN to estimate the unknown blur function. Recently, a kernel-free end-to-end approaches by Noorozi~\cite{Noroozi2017MotionWild} and Nah~\cite{Nah2016DeepDeblurring} that uses multi-scale CNN to directly deblur the image. Ramakrishnan~\etal~\cite{DeepGF} use the combination of pix2pix framework~\cite{pix2pix} and densely connected convolutional networks~\cite{DenseCNN} to perform blind kernel-free image deblurring. Such methods are able to deal with different sources of the blur.

\subsection{Generative adversarial networks}
The idea of generative adversarial networks, introduced by Goodfellow~\etal\cite{GAN}, is to define a game between two competing networks: the discriminator and the generator. The generator receives noise as an input and generates a sample. A discriminator receives a real and generated sample and is trying to distinguish between them. The goal of the generator is to fool the discriminator by generating perceptually convincing samples that can not be distinguished from the real one.  
The game between the generator $G$ and discriminator $D$ is the minimax objective:
\begin{equation}
\displaystyle \min_G \max_D\mathop{\mathbb{E}}_{x\backsim \mathbb{P}_r} [\text{log}(D(x))] + \mathop{\mathbb{E}}_{\tilde{x}\backsim \mathbb{P}_g} [\text{log}(1 - D(\tilde{x}))]
\end{equation}
where $\mathbb{P}_r$ is the data distribution and $\mathbb{P}_g$ is the model distribution, defined by $\tilde{x} = G(z), z\backsim P(z)$, the input $z$ is a sample from a simple noise distribution.
GANs are known for its ability to generate samples of good perceptual quality, however, training of vanilla version suffer from many problems such as mode collapse, vanishing gradients etc, as described in~\cite{GAN++}. Minimizing the value function in GAN is equal to minimizing the Jensen-Shannon divergence between the data and model distributions on $x$.
Arjovsky~\etal~\cite{WGAN} discuss the difficulties in GAN training caused by JS divergence approximation and propose to use the Earth-Mover (also called Wasserstein-1) distance $W(q,p)$. The value function for WGAN is constructed using Kantorovich-Rubinstein duality~\cite{KRDuality}:
\begin{equation}
\label{eq:1}
\displaystyle \min_G \max_{D \in \mathcal{D}}\mathop{\mathbb{E}}_{x\backsim \mathbb{P}_r} [D(x)] - \mathop{\mathbb{E}}_{\tilde{x}\backsim \mathbb{P}_g} [D(\tilde{x})]
\end{equation}
where $\mathcal{D}$ is the set of $1-$Lipschitz functions and $\mathbb{P}_g$ is once again the model distribution
The idea here is that critic value approximates $K \cdot W(P_r,P_\theta)$, where $K$ is a Lipschitz constant and $W(P_r,P_\theta)$ is a Wasserstein distance. In this setting, a discriminator network is called critic and it approximates the distance between the samples. To enforce Lipschitz constraint in WGAN Arjovsky~\etal add weight clipping to $[-c,c]$. Gulrajani~\etal~\cite{WGAN-GP} propose to add a gradient penalty term instead:
\begin{equation}
\lambda\displaystyle \mathop{\mathbb{E}}_{\tilde{x}\backsim \mathbb{P}_{\tilde{x}}} [(\|\nabla_{\tilde{x}}D(\tilde{x})\|_2 - 1)^2]
\end{equation}
to the value function as an alternative way to enforce the Lipschitz constraint. This approach is robust to the choice of generator architecture and requires almost no hyperparameter tuning. This is crucial for image deblurring as it allows to use novel lightweight neural network architectures in contrast to standard Deep ResNet architectures, previously used for image deblurring~\cite{Nah2016DeepDeblurring}. 
\subsection{Conditional adversarial networks}
	Generative Adversarial Networks have been applied to different image-to-image translation problems, such as super resolution~\cite{SRGAN}, style transfer~\cite{StyleTransferGAN}, product photo generation ~\cite{PhotoGAN} and others. Isola~\etal~\cite{pix2pix} provides a detailed overview of those approaches and present conditional GAN architecture also known as \emph{pix2pix}. Unlike vanilla GAN, cGAN learns a mapping from observed image $x$ and random noise vector $z$, to $y: G : {x, z} \to y$. Isola~\etal also put a condition on the discriminator and use U-net architecture~\cite{U-Net} for generator and Markovian discriminator which allows achieving perceptually superior results on many tasks, including synthesizing photos from label maps, reconstructing objects from edge maps, and colorizing images.
\section{The proposed method}
\label{s:method}
 The goal is to recover sharp image $I_S$ given only a blurred image $I_B$ as an input, so no information about the blur kernel is provided. Debluring is done by the trained CNN $G_{\theta_G}$, to which we refer as the Generator. For each $I_B$ it estimates corresponding $I_S$ image. In addition, during the training phase, we introduce critic the network $D_{\theta_D}$ and train both networks in an adversarial manner.
\begin{figure}[htb]
\centering
 \includegraphics[width=0.95\linewidth]{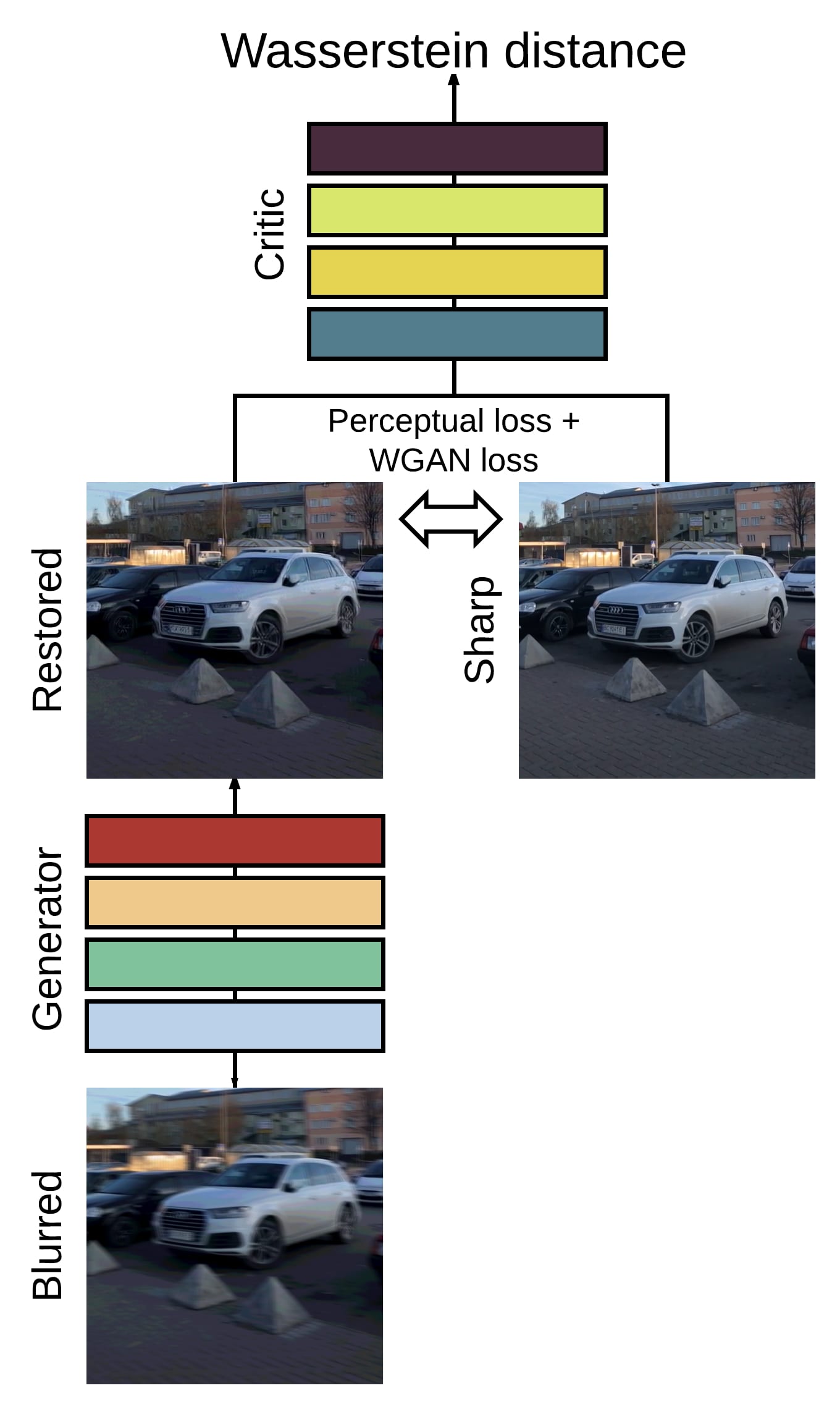}
   \caption{DeblurGAN training. The generator network takes the blurred image as input and produces the estimate of the sharp image. The critic network takes the restored and sharp images and outputs a distance between them. The total loss consists of the WGAN loss from critic and the perceptual loss~\cite{Johnson2016Perceptual}. The perceptual loss is the difference between the VGG-19~\cite{VGGNet2014} \emph{conv3.3} feature maps of the sharp and restored images. At test time, only the generator is kept.}
 \label{fig:GAN-arch}
\end{figure}
\subsection{Loss function}
 \begin{figure*}[htb]
	\centering
        \includegraphics[width=0.21\textwidth]{1351sharp}
        \includegraphics[width=0.28\textwidth]{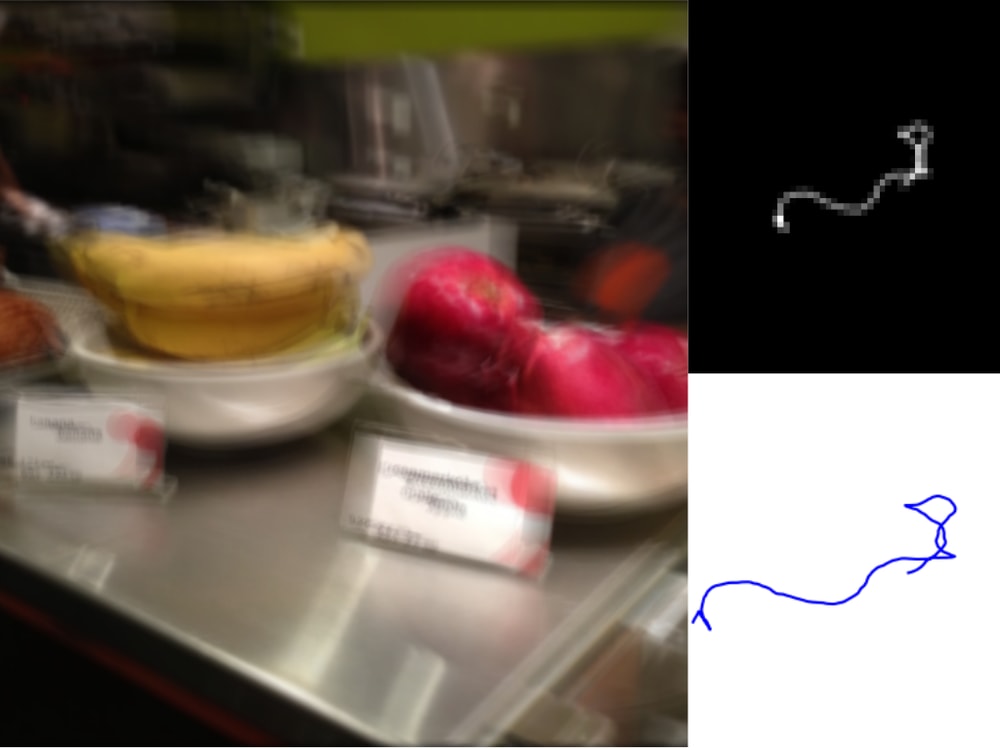}
        \includegraphics[width=0.21\textwidth]{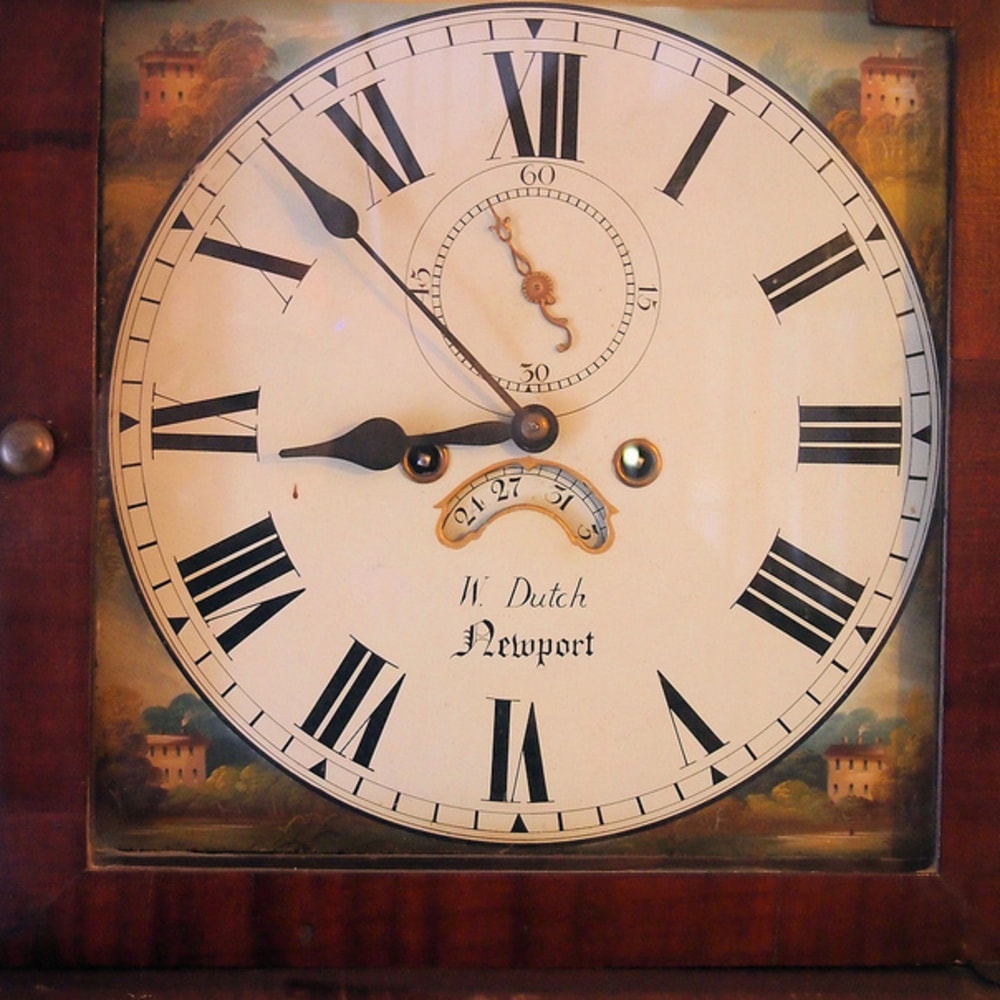}
        \includegraphics[width=0.28\textwidth]{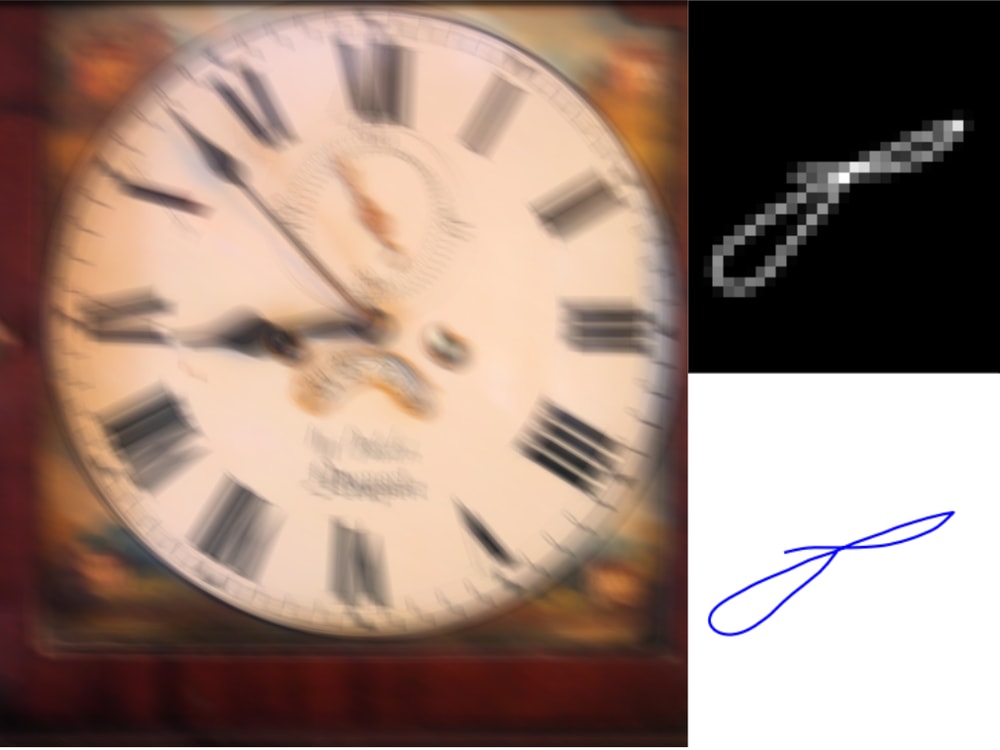}
    \caption{Examples of generated camera motion trajectory and the blur kernel and the corresponding blurred images.}\label{fig:process}
\end{figure*}
   We formulate the loss function as a combination of content and adversarial loss:
    \begin{equation}
    \mathcal{L} =\underbrace{\underbrace{\mathcal{L}_{GAN}}_{adv\:loss} \quad + \quad\underbrace{\lambda \cdot\mathcal{L}_{X}}_{content\:loss} }_{total\:loss}
    \end{equation}
where the $\lambda$ equals to 100 in all experiments. Unlike Isola~\etal\cite{pix2pix} we do not condition the discriminator as we do not need to penalize mismatch between the input and output. 
 \textbf{Adversarial loss}
	Most of the papers related to conditional GANs, use vanilla GAN objective as the loss~\cite{SRGAN}\cite{Nah2016DeepDeblurring} function. Recently~\cite{CycleGAN2017} provides an alternative way of using least aquare GAN~\cite{LSGAN} which is more stable and generates higher quality results. We use WGAN-GP~\cite{WGAN-GP} as the critic function, which is shown to be robust to the choice of generator architecture~\cite{WGAN}. Our premilinary experiments with different architectures confirmed that findings and we are able to use architecture much lighter than ResNet152~\cite{Nah2016DeepDeblurring}, see next subsection.
    The loss is calculated as the following:
 \begin{equation}
 \mathcal{L}_{GAN} = \displaystyle \mathop\sum_{n=1}^{N}-D_{\theta_D}(G_{\theta_G}(I^B))
 \end{equation}
 DeblurGAN trained without GAN component converges, but produces smooth and blurry images.
 
  \textbf{Content loss}. Two classical choices for "content" loss function are \textbf{L1} or \textbf{MAE} loss, \textbf{L2} or \textbf{MSE} loss on raw pixels. Using those functions as sole optimization target leads to the blurry artifacts on generated images due to the pixel-wise average of possible solutions in the pixel space~\cite{SRGAN}. Instead, we adopted recently proposed Perceptual loss~\cite{Johnson2016Perceptual}. Perceptual loss is a simple L2-loss, but based on the difference of the generated and target image CNN feature maps. It is defined as following:
 \begin{center}
	$\mathcal{L}_{X} = \frac{1}{W_{i,j}H_{i,j}}\displaystyle \mathop\sum_{x=1}^{W_{i,j}}\displaystyle \mathop\sum_{y=1}^{H_{i,j}}(\phi_{i,j}(I^S)_{x,y} - \phi_{i,j}(G_{\theta_G}(I^B))_{x,y})^2$
 \end{center}
 where $\phi_{i,j}$ is the feature map obtained by the j-th convolution (after activation) before the i-th maxpooling layer within the VGG19 network, pretrained on ImageNet~\cite{ImageNet}, $W_{i,j}$ and $H_{i,j}$ are the dimensions of the feature maps. In our work we use activations from $VGG_{3,3}$ convolutional layer. The activations of the deeper layers represents the features of a higher abstraction~\cite{FergusCNN}\cite{SRGAN}. The perceptual loss focuses on restoring general content~\cite{pix2pix}~\cite{SRGAN} while adversarial loss focuses on restoring texture details.
 DeblurGAN trained without Perceptual loss or with simple MSE on pixels instead doesn't converge to meaningful state.
 
 \textbf{Additional regularization.}
  We have also tried to add TV regularization and model trained with it yields worse performance -- 27.9 vs. 28.7 w/o PSNR on GoPro dataset.
  \subsection{Network architecture}
  Generator CNN architecture is shown in Figure~\ref{fig:arch}. It is similar to one proposed by Johnson~\etal\cite{Johnson2016Perceptual} for style transfer task. It contains two strided convolution blocks with stride $\frac{1}{2}$, nine residual blocks~\cite{ResNet} (ResBlocks) and two transposed convolution blocks. Each ResBlock consists of a convolution layer, instance normalization layer~\cite{InstanceNorm}, and ReLU~\cite{ReLU2010} activation. Dropout~\cite{Dropout} regularization with a probability of 0.5 is added after the first convolution layer in each ResBlock. 
  In addition, we introduce the global skip connection which we refer to as ResOut. CNN learns a residual correction $I_R$ to the blurred image $I_B$, so $I_S = I_B + I_R$. 
We find that such formulation makes training faster and resulting model generalizes better.
	During the training phase, we define a critic network $D_{\theta_D}$, which is Wasserstein GAN~\cite{WGAN} with gradient penalty~\cite{WGAN-GP}, to which we refer as WGAN-GP. The architecture of critic network is identical to PatchGAN ~\cite{pix2pix,StyleTransferGAN}. All the convolutional layers except the last are followed by InstanceNorm layer and LeakyReLU~\cite{LeakyRelu2015} with $\alpha = 0.2$.
%
\section{Motion blur generation}
 \begin{figure}[thb]
  \includegraphics[width=0.5\textwidth]{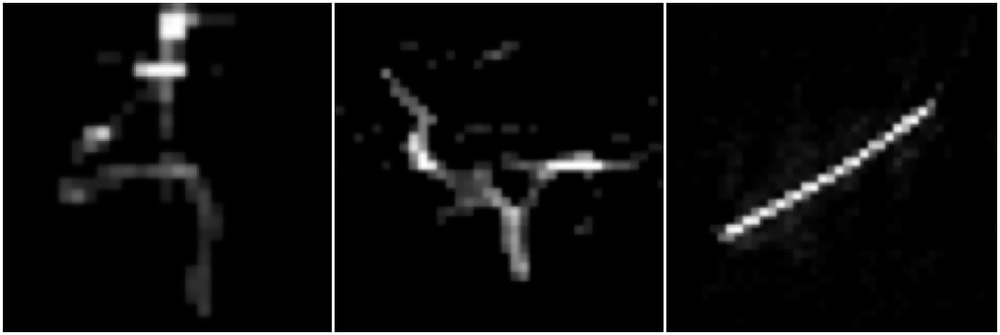} \\
  \includegraphics[width=0.5\textwidth]{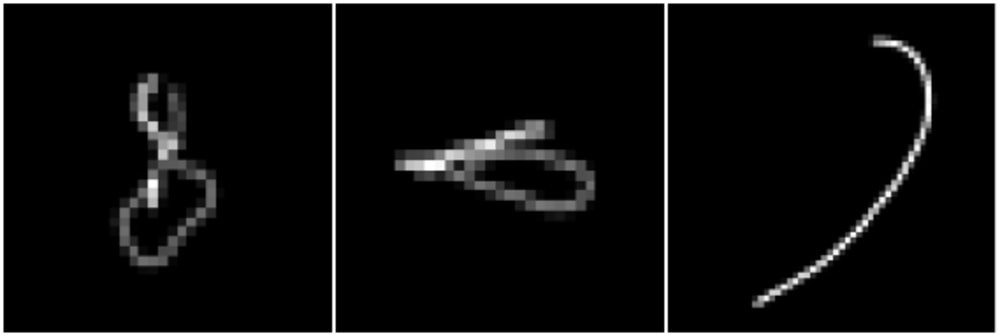}
  \caption{Top row: Blur kernels from real-world images estimated by Fergus~\etal\cite{Fergus}. Bottom row: Synthetically generated kernels by our method. Our randomized method can simulate wide variety of realistic blur kernels with different level of non-linearity.}
  \label{fig:kernels}
\end{figure}
	There is no easy method to obtain image pairs of corresponding sharp and blurred images for training.
    A typical approach to obtain image pairs for training is to use a high frame-rate camera to simulate blur using average of sharp frames from video~\cite{Noroozi2017MotionWild,Nah2016DeepDeblurring}. It allows to create realistic blurred images but limits the image space only to scenes present in taken videos and makes it complicated to scale the dataset. 
    Sun~\etal\cite{SunLearningRemoval} creates synthetically blurred images by convolving clean natural images with one out of 73 possible linear motion kernels, Xu~\etal\cite{Xu} also use linear motion kernels to create synthetically blurred images. Chakrabarti ~\cite{Chakrabarti2016ADeblurring} creates blur kernel by sampling 6 random points and fitting a spline to them. 
    We take a step further and propose a method, which simulates more realistic and complex blur kernels.
	We follow the idea described by Boracchi and Foi~\cite{boracchiFoiTIP12a} of random trajectories generation. Then the kernels are generated by applying sub-pixel interpolation to the trajectory vector. Each trajectory vector is a complex valued vector, which corresponds to the discrete positions of an object following 2D random motion in a continuous domain. 
    Trajectory generation is done by Markov process, summarized in Algorithm~\ref{alg:thealg}. Position of the next point of the trajectory is randomly generated based on the previous point velocity and position, gaussian perturbation, impulse perturbation and deterministic inertial component.
\begin{algorithm}
\caption{\textbf{Motion blur kernel generation.}\\  Parameters:\\
$M$ = 2000 -- number of iterations,\\
$L_{max}$ = 60 -- max length of the movement, \\
$p_s$ = 0.001 -- probability of impulsive shake,\\
$I$ -- inertia term, uniform from (0,0.7),\\
$p_b$ -- probability of big shake, uniform from (0,0.2),\\
$p_g$ -- probability of gaussian shake, uniform from (0,0.7),\\
$\phi$ -- initial angle, uniform from (0,2$\pi$),\\
$x$ -- trajectory vector.}
\label{alg:thealg}
\begin{algorithmic}[1]
\Procedure{Blur}{Img, $M,  L_{max}, p_s$}
\State $v_0\gets cos(\phi) + sin(\phi)*i$
\State $v \gets v_o * L_{max} / (M-1)$
\State $x = \text{zeros}(M, 1)$
\For{$t = 1$ to ${M - 1}$}
\If {$\text{randn} < p_b * p_s$}
\State $\text{nextDir}\gets  2 \cdot v\cdot  e^{ i * (\pi + (\text{randn} - 0.5))})$
\Else:
\State $\text{nextDir}\gets 0 $
\EndIf
\State $dv \gets \text{nextDir} + p_s * (p_g * (\text{randn} + i * \text{randn})*I * x[t] * (L_{max} / (M - 1))$
\State $v \gets v + dv$
\State $v \gets (v / abs(v)) * L_{max} / (M - 1)$
\State $x[t + 1] \gets x[t] + v$
\EndFor
\State $\text{Kernel} \gets \text{sub pixel interpolation}(x)$
\State $\text{Blurred image} \gets conv(\text{Kernel}, \text{Img})$
\State \textbf{return} $\text{Blurred image}$
\EndProcedure
\end{algorithmic}
\end{algorithm}
\section{Training Details}
\begin{figure*}[htb]
	\centering
        \includegraphics[width=0.33\textwidth]{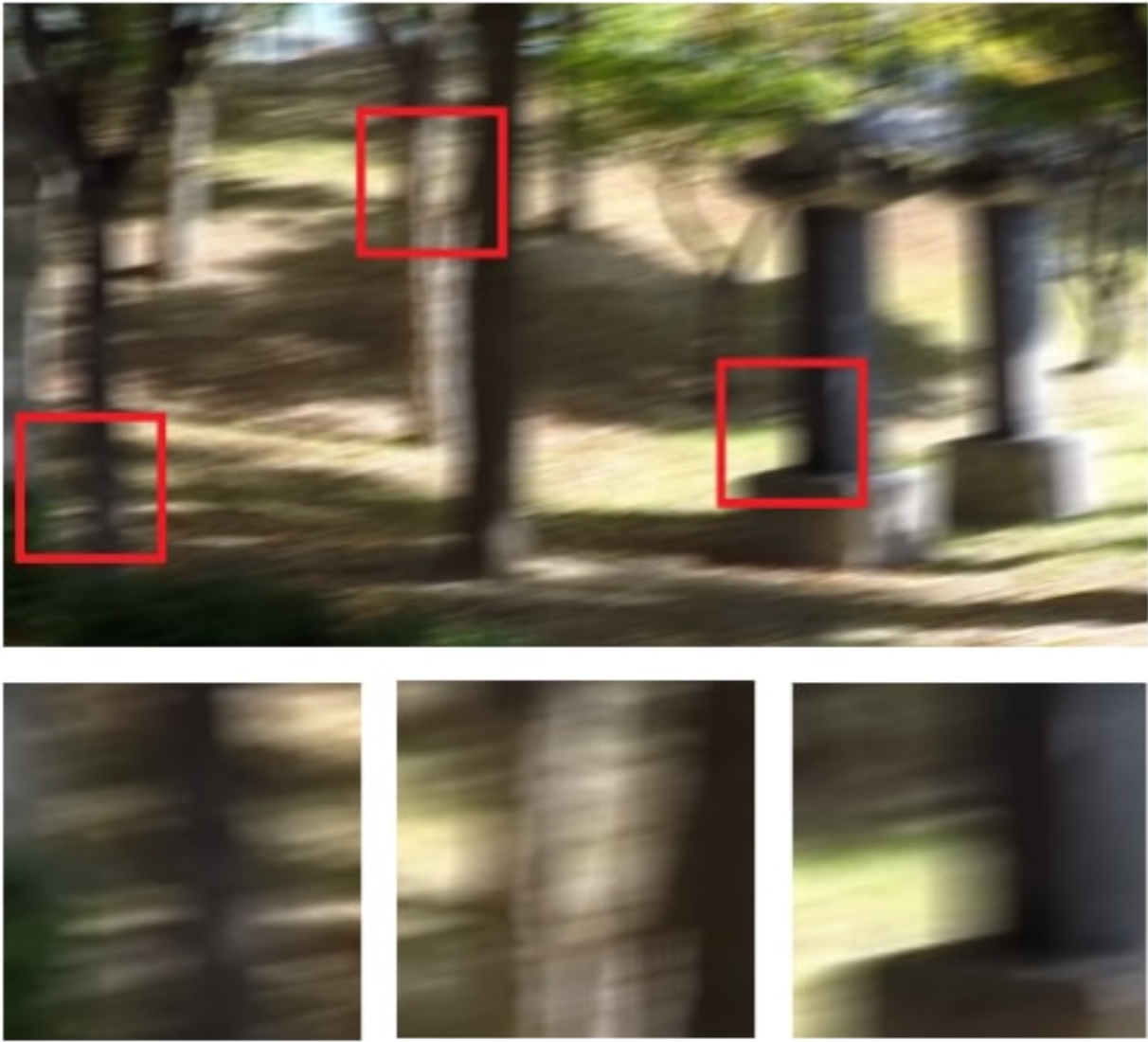}
        \includegraphics[width=0.33\textwidth]{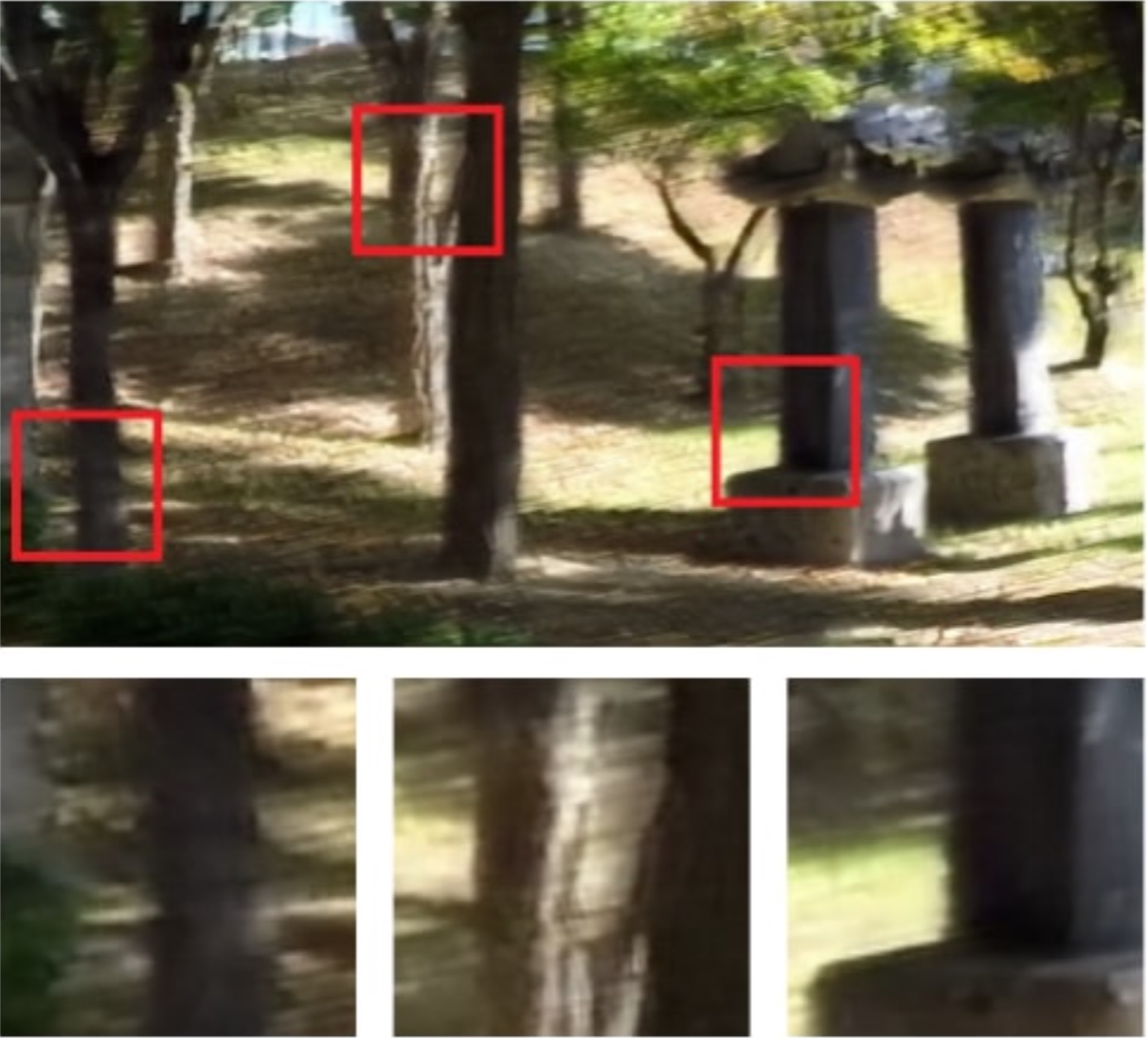}
        \includegraphics[width=0.33\textwidth]{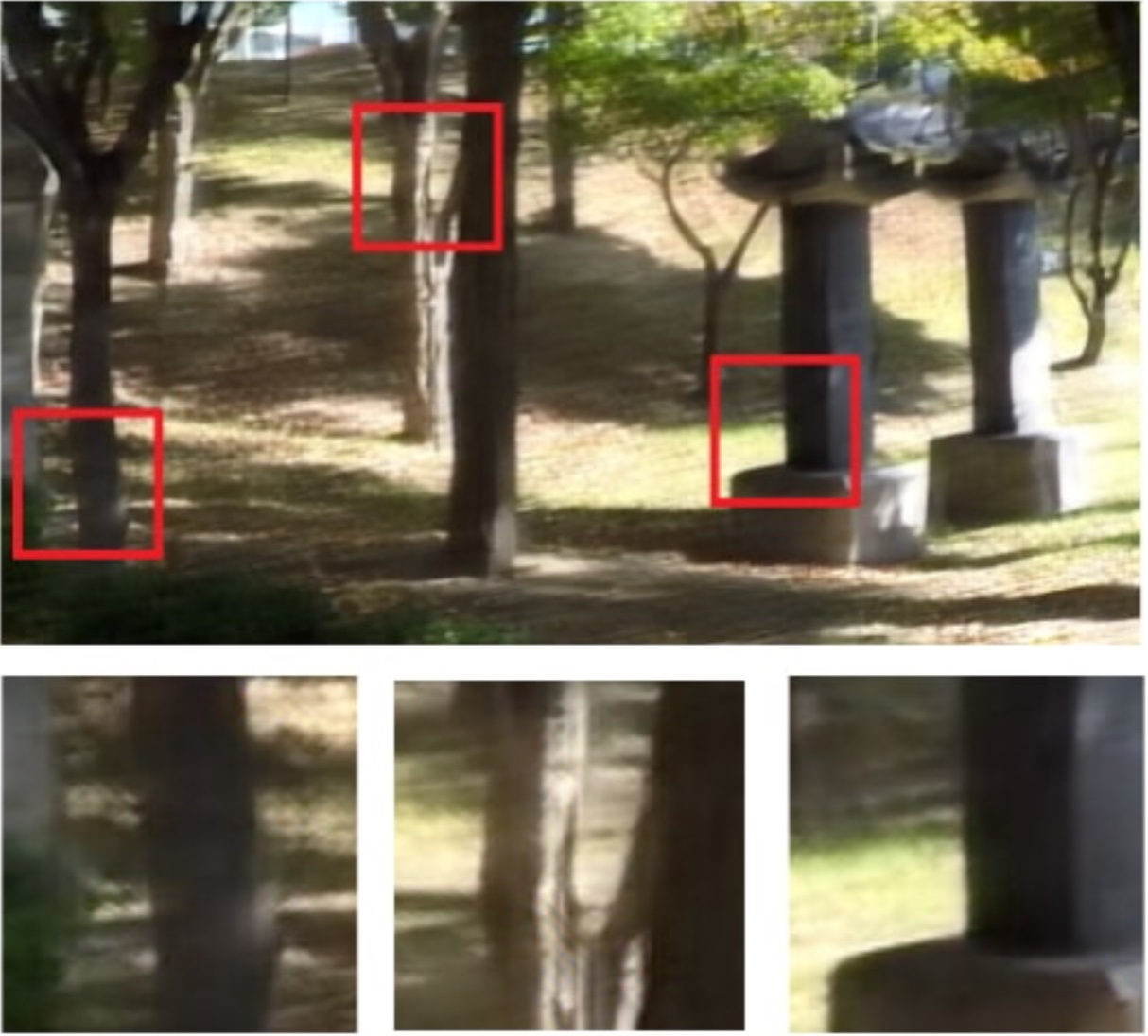}
        \includegraphics[width=0.33\textwidth]{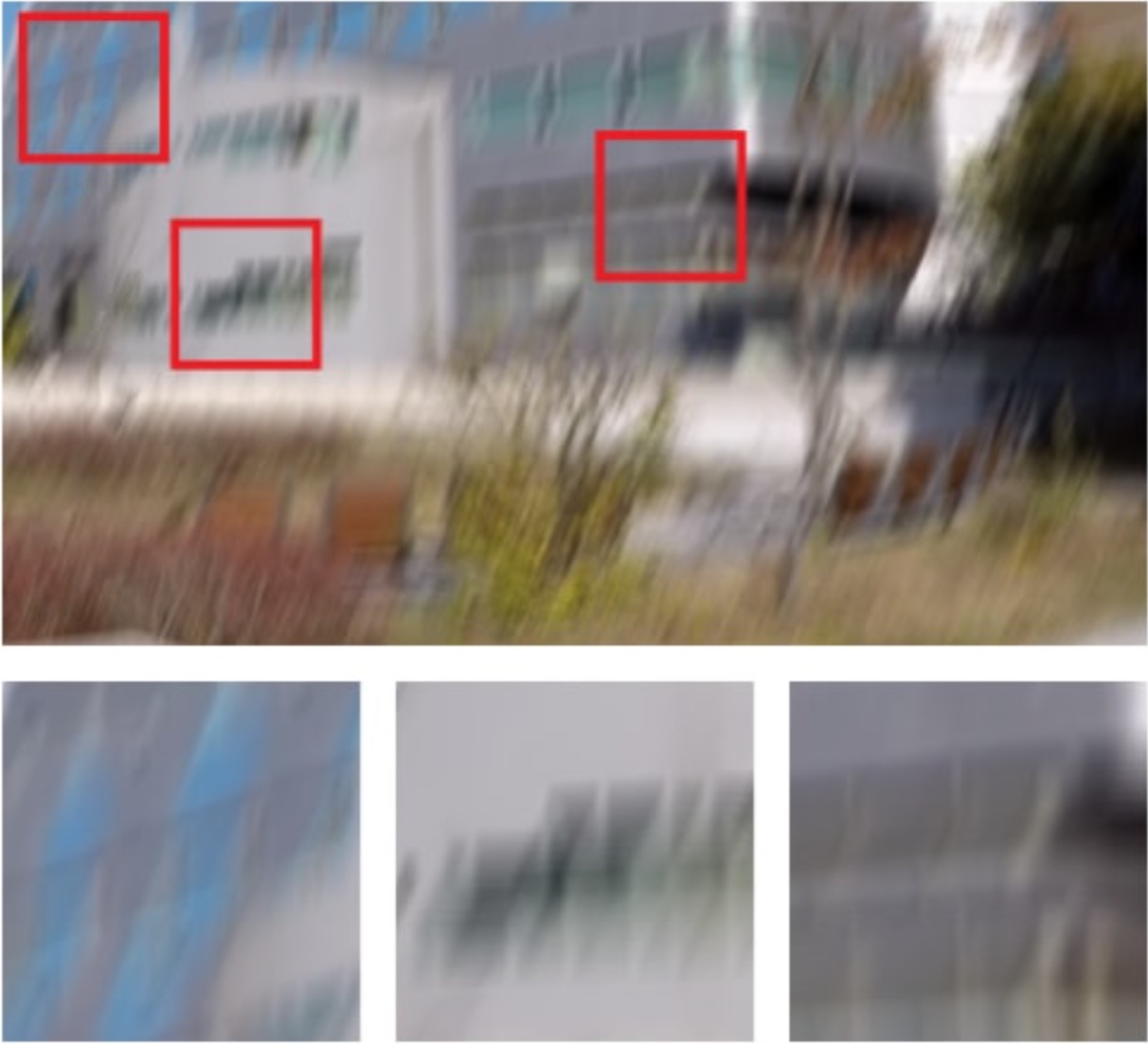}
        \includegraphics[width=0.33\textwidth]{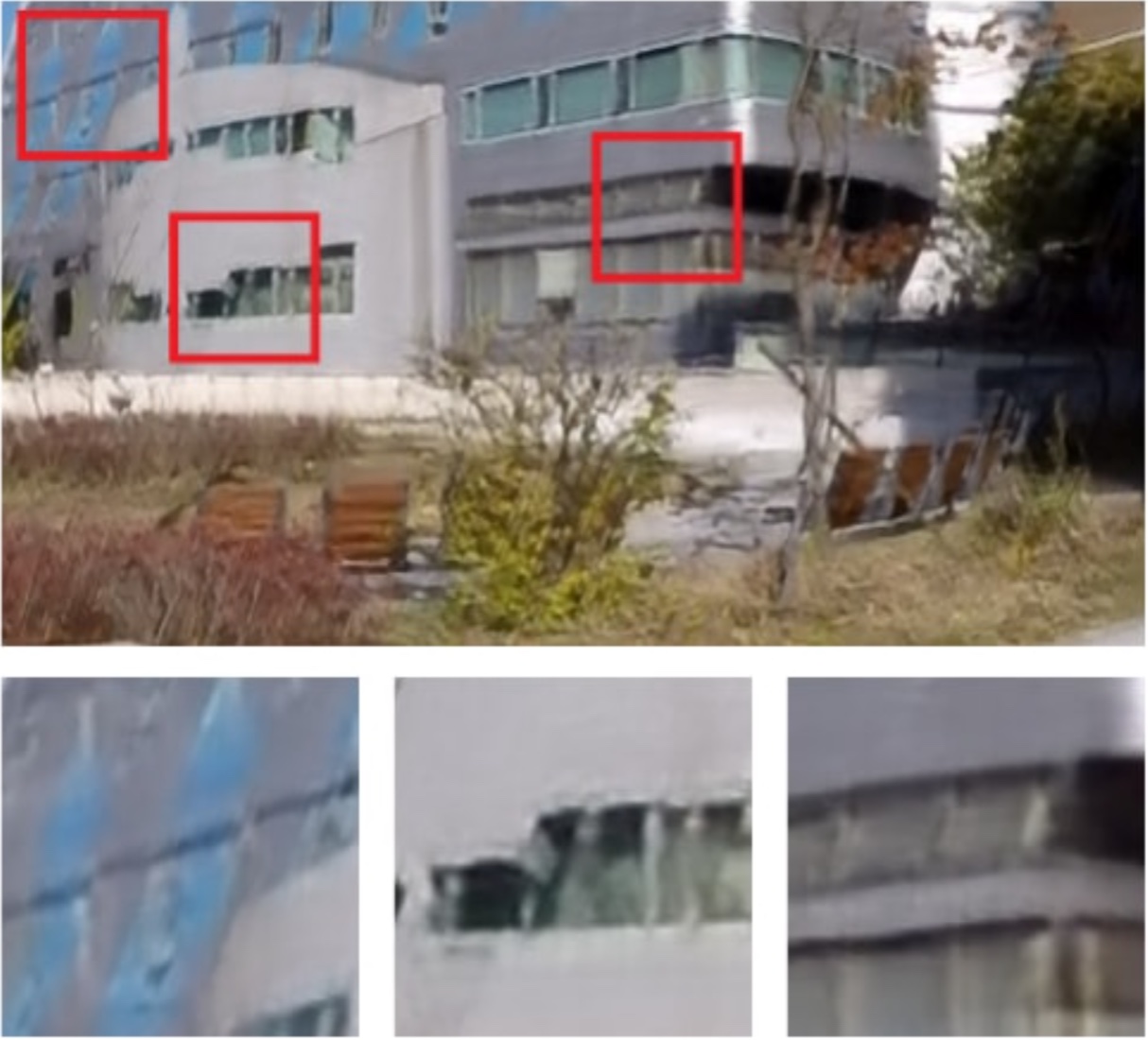}
        \includegraphics[width=0.33\textwidth]{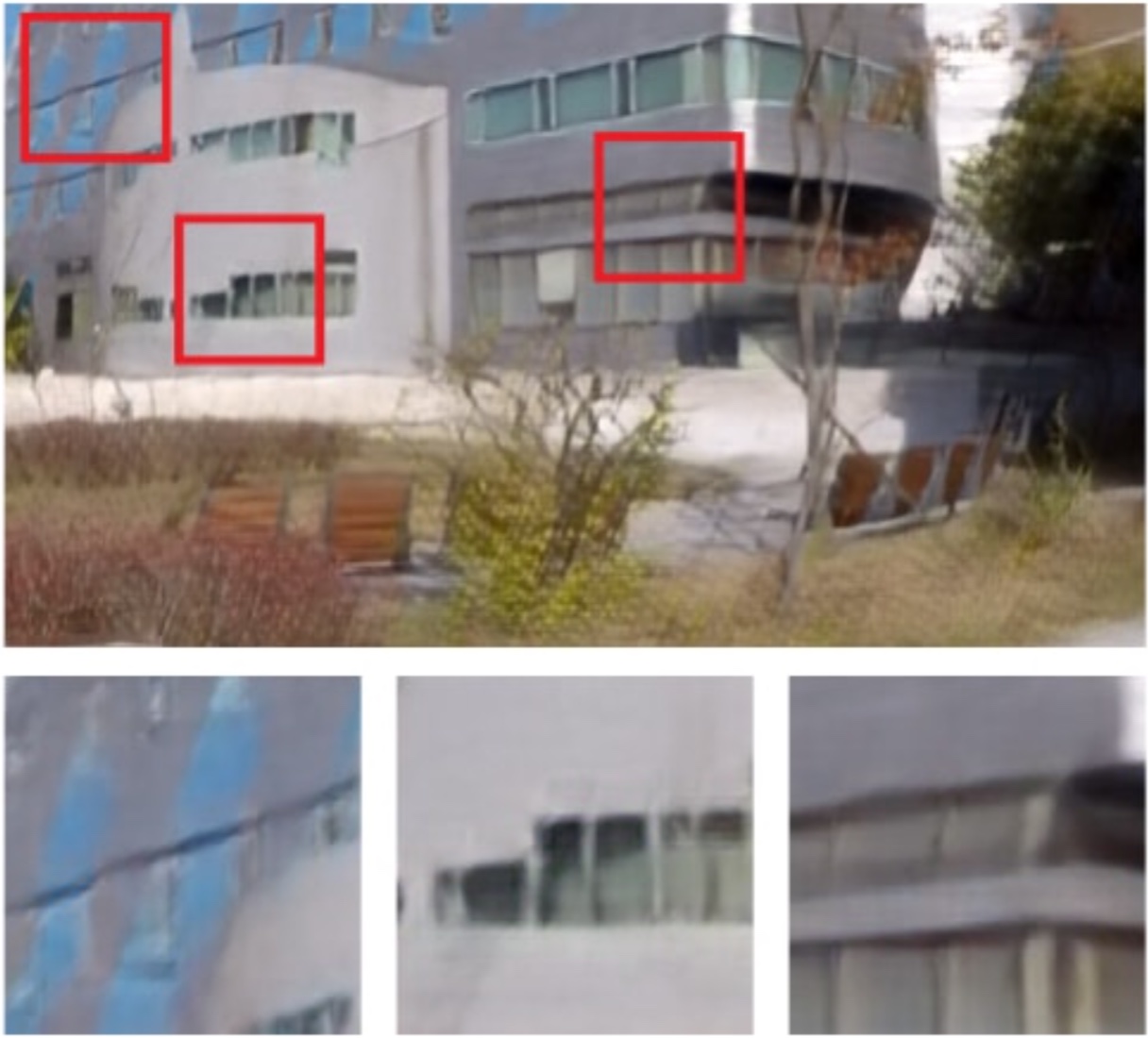}
    \caption{Results on the GoPro test dataset. From left to right: blurred photo, Nah~\etal~\cite{Nah2016DeepDeblurring}, DeblurGAN.}
    \label{fig:gopro_test}
\end{figure*}
\begin{figure*}[htb]
	\centering
        \includegraphics[width=0.33\textwidth]{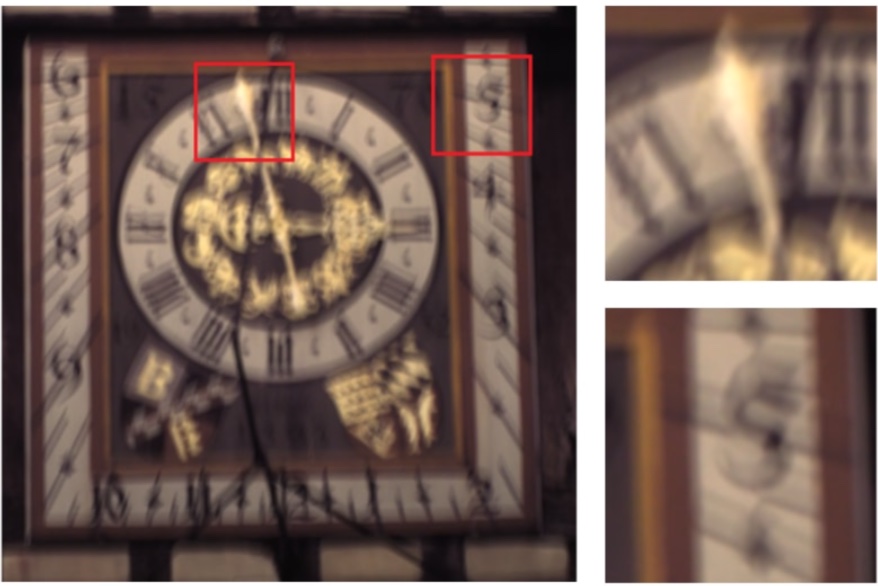}
        \includegraphics[width=0.33\textwidth]{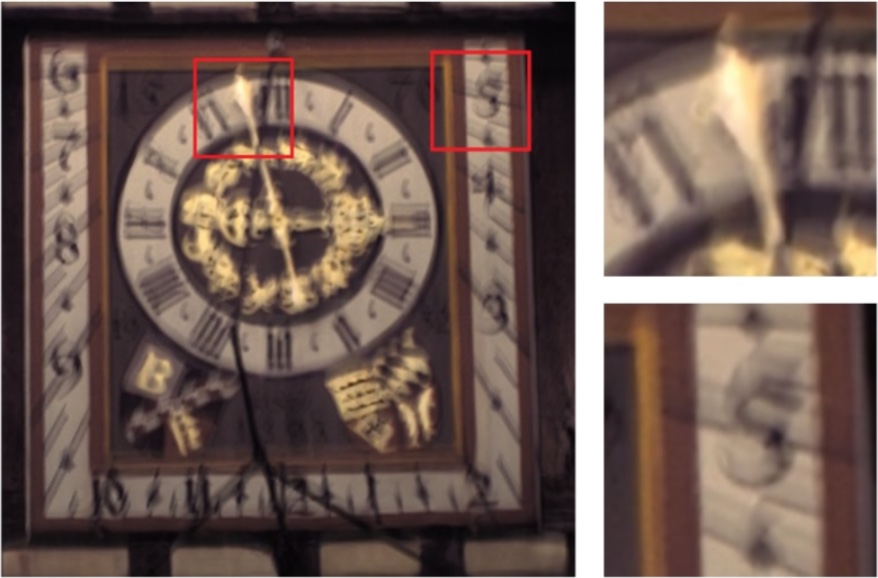}
        \includegraphics[width=0.33\textwidth]{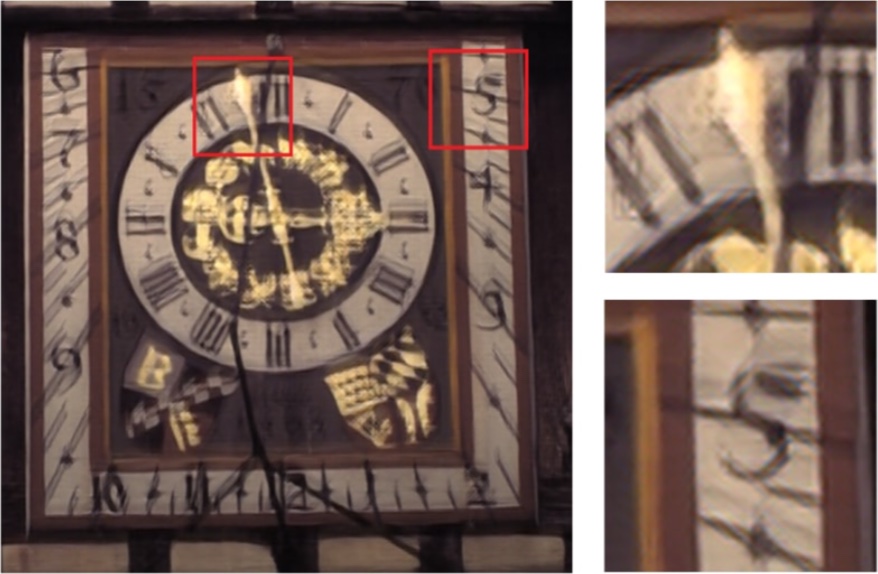}
    \vspace{1em}
        \includegraphics[width=0.33\textwidth]{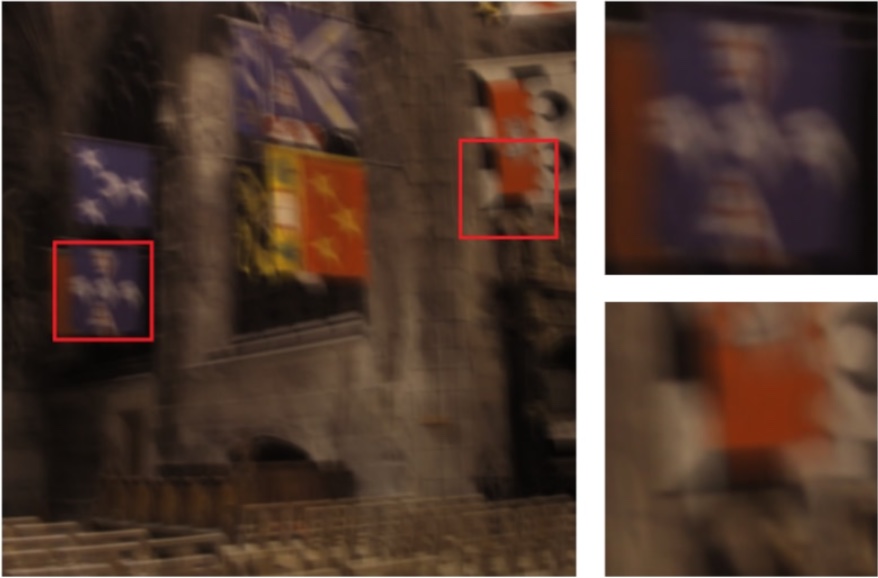}
        \includegraphics[width=0.33\textwidth]{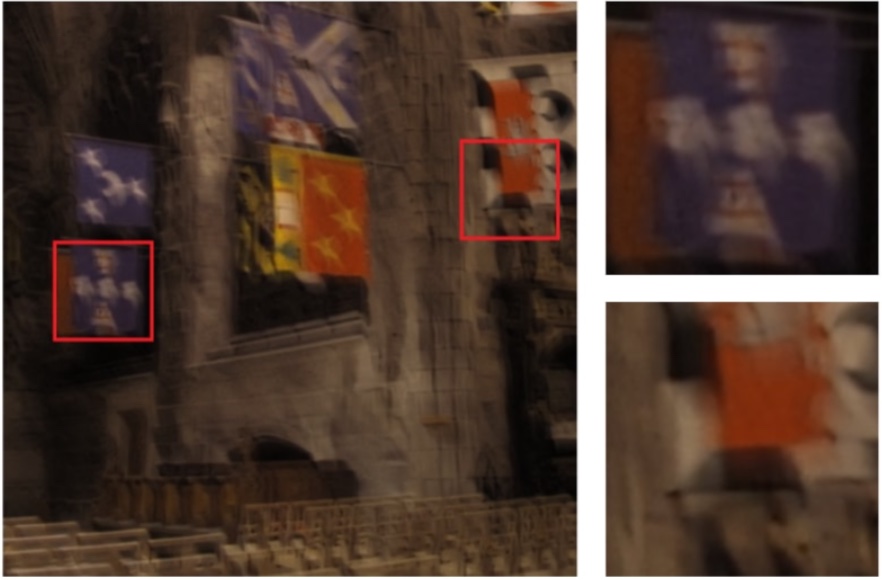}
        \includegraphics[width=0.33\textwidth]{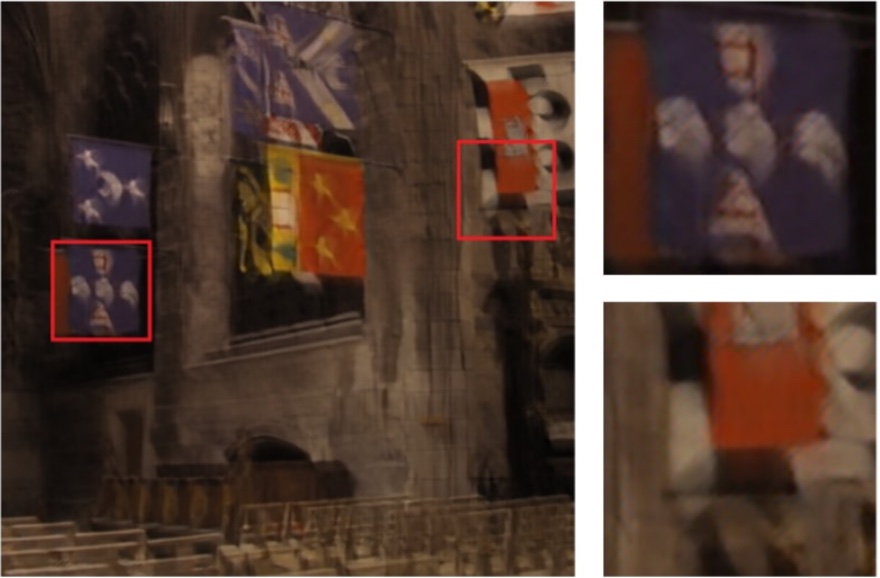}
    \caption{Results on the Kohler dataset. From left to right: blurred photo, Nah~\etal\cite{Nah2016DeepDeblurring}, DeblurGAN.}
    \label{fig:kohler_test}
\end{figure*}
	We implemented all of our models using PyTorch\cite{pytorch} deep learning framework. The training was performed on a single Maxwell GTX Titan-X GPU using three different datasets. The first model to which we refer as \emph{DeblurGAN\textsubscript{WILD}} was trained on a random crops of size 256x256 from 1000 GoPro training dataset images~\cite{Nah2016DeepDeblurring} downscaled by a factor of two. The second one \emph{DeblurGAN\textsubscript{Synth}} was trained on 256x256 patches from MS COCO dataset blurred by method, presented in previous Section. 
    We also trained \emph{DeblurGAN\textsubscript{Comb}} on a combination of synthetically blurred images and images taken in the wild, where the ratio of synthetically generated images to the images taken by a high frame-rate camera is 2:1. As the models are fully convolutional and are trained on image patches they can be applied to images of arbitrary size. 
    For optimization we follow the approach of ~\cite{WGAN} and perform 5 gradient descent
steps on $D_{\theta_D}$, then one step on $G_{\theta_G}$, using Adam ~\cite{ADAM} as a solver. The learning rate is set initially to $10^{-4}$ for both generator and critic. After the first $150$ epochs we linearly decay the rate to zero over the next $150$ epochs. At inference time we follow the idea of~\cite{pix2pix} and apply both dropout and instance normalization. All the models were trained with a batch size = 1, which showed empirically better results on validation. The training phase took 6 days for training one \emph{DeblurGAN} network.
\section{Experimental evaluation}
\label{s:results}
\subsection{GoPro Dataset}
\begin{table}[htb]
\small
\caption{Peak signal-to-noise ratio and the structural similarity measure, mean over the GoPro test dataset of 1111 images. All models were tested on the \emph{linear} image subset. State-of-art results $(*)$ by Nah~\etal~\cite{Nah2016DeepDeblurring} obtained on the \emph{gamma} subset.}
\label{T:gopro}
\centering
\ra{1.0}
\setlength\tabcolsep{3pt}
\begin{tabular}{ccccccc}
\toprule
 & Sun~\etal&Nah~\etal & Xu~\etal & \multicolumn{3}{c}{DeblurGAN} \\
\cmidrule(r){5-7}
Metric& \cite{SunLearningRemoval}& \cite{Nah2016DeepDeblurring}& \cite{XuUnnaturalDeblurring} &  \textit{WILD} & \textit{Synth} & \textit{Comb} \\
\cmidrule(r){1-7}
PSNR & 24.6 &  28.3/\textbf{29.1*}  & 25.1& 27.2  & 23.6 & \textbf{28.7} \\
SSIM & 0.842  &  0.916 &0.89 & 0.954 & 0.884 & \textbf{0.958}\\
\cmidrule(r){1-7}
Time & 20 min  & 4.33 s & 13.41 s & \multicolumn{3}{ c }{\textbf{0.85 s}} \\
\bottomrule
\end{tabular}
\end{table}
	GoPro dataset\cite{Nah2016DeepDeblurring} consists of 2103 pairs of blurred and sharp images in 720p quality, taken from various scenes. We compare the results of our models with state of the art models~\cite{SunLearningRemoval}, \cite{Nah2016DeepDeblurring} on standard metrics and also show the running time of each algorithm on a single \textbf{GPU}. Results are in Table\ref{T:gopro}. DeblurGAN shows superior results in terms of structured self-similarity, is close to state-of-the-art in peak signal-to-noise-ratio and provides better looking results by visual inspection. In contrast to other neural models, our network does not use L2 distance in pixel space so it is not directly optimized for PSNR metric. 
    It can handle blur caused by camera shake and object movement, does not suffer from usual artifacts in kernel estimation methods and at the same time has more than 6x fewer parameters comparing to Multi-scale CNN , which heavily speeds up the inference. Deblured images from test on GoPro dataset are shown in Figure~\ref{fig:gopro_test}. 
\subsection{Kohler dataset}
	Kohler dataset ~\cite{Kohler} consists of 4 images blurred with 12 different kernels for each of them. This is a standard benchmark dataset for evaluation of blind deblurring algorithms. The dataset is generated by recording and analyzing real camera motion, which is played back on a robot platform such that a sequence of sharp images is recorded sampling the 6D camera motion trajectory.
    Results are in Table~\ref{T:kohler}, similar to GoPro evaluation.
\begin{table*}[htb]
\small
\caption{Peak signal-to-noise ratio and structural similarity measure, mean on the Kohler dataset. Xu~\etal\cite{XuUnnaturalDeblurring} and Whyte~\etal\cite{whyte} are non-CNN blind deblurring methods, whereas Sun~\etal\cite{SunLearningRemoval} and Nah~\etal\cite{Nah2016DeepDeblurring} use CNN.}
\label{T:kohler}
\centering
\begin{tabular}{cccccccc}
\toprule
Method & Sun~\etal & Nah~\etal & Xu~\etal & Whyte~\etal & \multicolumn{3}{ c }{DeblurGAN}  \\ 
\cmidrule(r){6-8}
Metric&\cite{SunLearningRemoval} & \cite{Nah2016DeepDeblurring} & \cite{XuUnnaturalDeblurring} & \cite{whyte} & \textit{WILD} & \textit{Synth} & \textit{Comb} \\
\cmidrule(r){1-8}
PSNR & 25.22 & 26.48 & \textbf{27.47} & 27.03 & 26.10 & 25.67 & 25.86 \\
SSIM & 0.773  & 0.807 & 0.811 & 0.809 & \textbf{0.816}  & 0.792 & 0.802 \\
\bottomrule
\end{tabular}
\end{table*}
\subsection{Object Detection benchmark on YOLO}
 \begin{figure*}[htb]
	\centering
    \begin{subfigure}[t]{0.24\textwidth}
        \includegraphics[width=\textwidth]{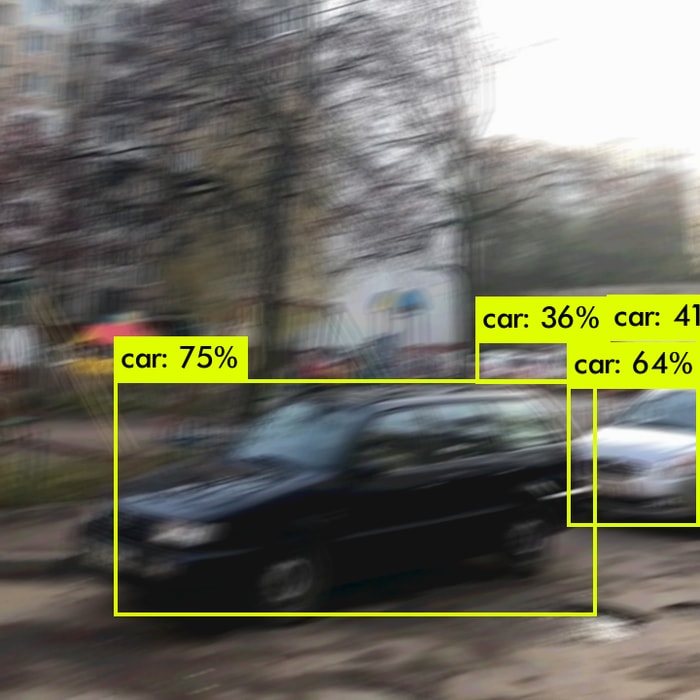}
        \caption{Blurred photo}
        \label{fig:yolo_blur}
    \end{subfigure}
    \begin{subfigure}[t]{0.24\textwidth}
        \includegraphics[width=\textwidth]{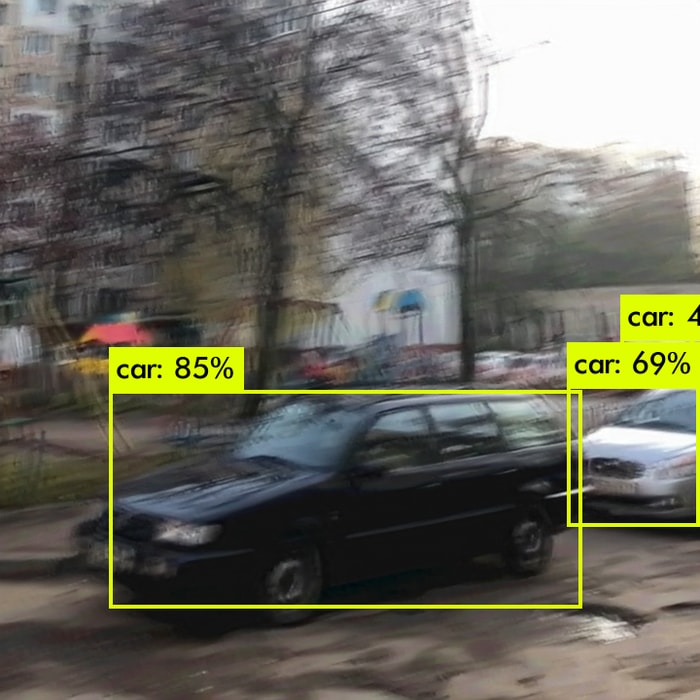}
        \caption{Nah~\etal~\cite{Nah2016DeepDeblurring}}
        \label{fig:yolo_kim}
    \end{subfigure}
    \begin{subfigure}[t]{0.24\textwidth}
        \includegraphics[width=\textwidth]{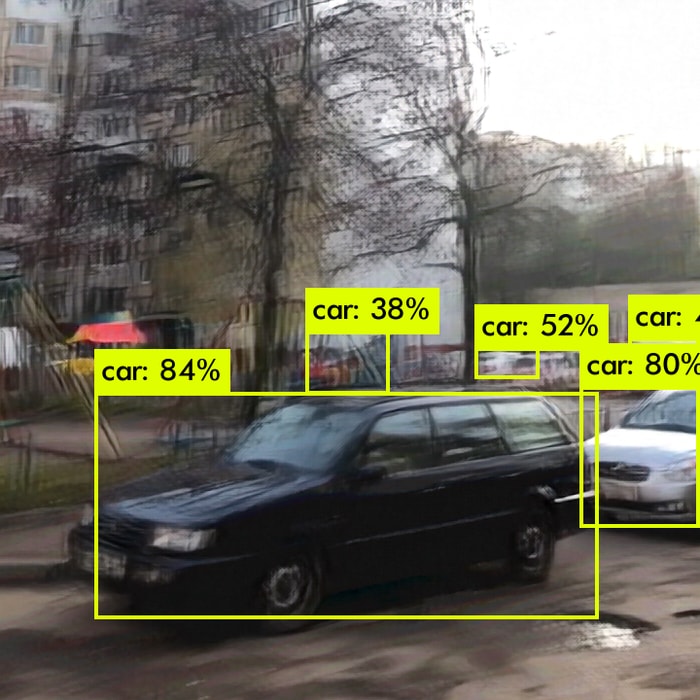}
        \caption{DeblurGAN}
        \label{fig:yolo_out}
    \end{subfigure}
    \begin{subfigure}[t]{0.24\textwidth}
        \includegraphics[width=\textwidth]{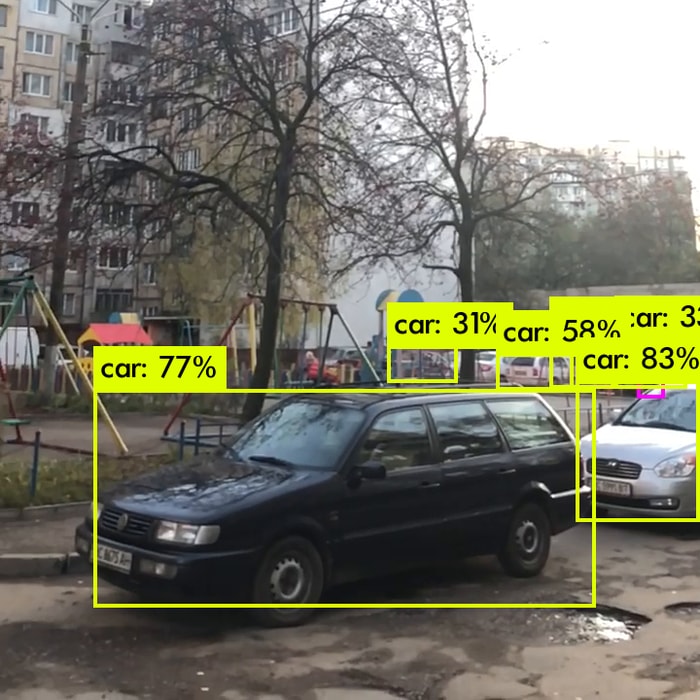}
        \caption{Sharp photo}
        \label{fig:yolo_sharp}
    \end{subfigure}
    \caption{YOLO object detection before and after deblurring}\label{fig:yolo}
\end{figure*}
	Object Detection is one of the most well-studied problems in computer vision with applications in different domains from autonomous driving to security. During the last few years approaches based on Deep Convolutional Neural Networks showed state of the art performance comparing to traditional methods. However, those networks are trained on limited datasets and in real-world settings images are often degraded by different artifacts, including motion blur, 
Similar to~\cite{Dehazing} and~\cite{EnhNet} we studied the influence of motion blur on object detection and propose a new way to evaluate the quality of deblurring algorithm based on results of object detection on a pretrained YOLO~\cite{YOLO} network. 

For this, we constructed a dataset of sharp and blurred street views by simulating camera shake using a high frame-rate video camera. Following~\cite{VideoBlur}\cite{Nah2016DeepDeblurring}\cite{Noroozi2017MotionWild} we take a random between 5 and 25 frames taken by 240fps camera and compute the blurred version of a middle frame as an average of those frames. All the frames are gamma-corrected with $\gamma=2.2$ and then the inverse function is taken to obtain the final blurred frame. Overall, the dataset consists of 410 pairs of blurred and sharp images, taken from the streets and parking places with different number and types of cars.

Blur source includes both camera shake and blur caused by car movement. The dataset and supplementary code are available online. 
Then sharp images are feed into the YOLO network and the result after visual verification is assigned as ground truth.
Then YOLO is run on blurred and recovered versions of images and average recall and precision between obtained results and ground truth are calculated. 
This approach corresponds to the quality of deblurring models on real-life problems and correlates with the visual quality and sharpness of the generated images, in contrast to standard PSNR metric. The precision, in general, is higher on blurry images as there are no sharp object boundaries and smaller object are not detected as it shown in Figure~\ref{fig:yolo}. 

Results are shown in Table~\ref{T:yolo}. DeblurGAN significantly outperforms competitors in terms of recall and F1 score.
\begin{table}[htb]
\ra{1}
\centering
\caption{Results of YOLO~\cite{YOLO} object detection on blurred and restored photos using DeblurGAN and Nah~\etal~\cite{Nah2016DeepDeblurring} algorithms. Results on corresponding sharp images are considered ground truth. DeblurGAN has higher recall and F1 score than its competitors.}
\label{T:yolo}
\begin{tabular}{cccc}
\toprule
Method & prec. & recall & F1 score \\
\cmidrule(r){1-4}
no deblur & 0.821 & 0.437 & 0.570 \\
Nah~\etal~\cite{Nah2016DeepDeblurring} &  \textbf{0.834} & 0.552 &  0.665\\
DeblurGAN WILD &0.764& 0.631& 0.691 \\
DeblurGAN synth & 0.801 & 0.517 & 0.628 \\
DeblurGAN comb & 0.671 & \textbf{0.742} & \textbf{0.704} \\
\bottomrule
\end{tabular}
%
\end{table}
\section{Conclusion}
	We described a kernel-free blind motion deblurring learning approach and introduced DeblurGAN which is a Conditional Adversarial Network that is optimized using a multi-component loss function. In addition to this, we implemented a new method for creating a realistic synthetic motion blur able to model different blur sources. 
    We introduce a new benchmark and evaluation protocol based on results of object detection and show that DeblurGAN significantly helps detection on blurred images.
    \section{Acknowledgements}
\label{sec:acknowledgement}
The authors were supported by the ELEKS Ltd., ARVI Lab, Czech Science Foundation Project GACR P103/12/G084, the Austrian Ministry for Transport, Innovation and Technology, the Federal Ministry of Science, Research and Economy, and the Province of Upper Austria in the frame of the COMET center, the CTU student grant SGS17/185/OHK3/3T/13. 
We thank Huaijin Chen for finding the bug in peak-signal-to-noise ratio evaluation. 
{\small
\bibliographystyle{ieee}
\bibliography{egbib}
}
\end{document}